\documentclass[runningheads]{llncs}

\usepackage{eccv}

\usepackage{eccvabbrv}

\usepackage{graphicx}
\usepackage{booktabs}

\usepackage{color}
\usepackage{colortbl}
\usepackage{multirow}
\usepackage{float} 

\usepackage[accsupp]{axessibility}  % Improves PDF readability for those with disabilities.

\usepackage{hyperref}

\usepackage{orcidlink}

\begin{document}

\title{On the Error Analysis of 3D Gaussian Splatting and an Optimal Projection Strategy} 

\titlerunning{On the Error Analysis of 3D-GS and an Optimal Projection Strategy}

\author{Letian Huang\orcidlink{0009-0003-1454-7824} \and
Jiayang Bai\orcidlink{0000-0001-6892-0338} \and
Jie Guo\thanks{Corresponding author.}\orcidlink{0000-0002-4176-7617} \and Yuanqi Li\orcidlink{0000-0003-4100-7471} \and Yanwen Guo\orcidlink{0000-0002-7605-5206}}

\authorrunning{L.~Huang et al.}

\institute{Nanjing University, Nanjing, China\\
\email{\{lthuang, jybai, yuanqili\}@smail.nju.edu.cn, \{guojie, ywguo\}@nju.edu.cn}\\
\url{https://letianhuang.github.io/op43dgs/}
}

\maketitle

\begin{abstract}
  3D Gaussian Splatting has garnered extensive attention and application in real-time neural rendering. Concurrently, concerns have been raised about the limitations of this technology in aspects such as point cloud storage, performance, and robustness in sparse viewpoints, leading to various improvements. However, there has been a notable lack of attention to the fundamental problem of projection errors introduced by the local affine approximation inherent in the splatting itself, and the consequential impact of these errors on the quality of photo-realistic rendering. This paper addresses the projection error function of 3D Gaussian Splatting, commencing with the residual error from the first-order Taylor expansion of the projection function. The analysis establishes a correlation between the error and the Gaussian mean position. Subsequently, leveraging function optimization theory, this paper analyzes the function's minima to provide an optimal projection strategy for Gaussian Splatting referred to Optimal Gaussian Splatting, which can accommodate a variety of camera models. Experimental validation further confirms that this projection methodology reduces artifacts, resulting in a more convincingly realistic rendering.
\end{abstract}

\begin{figure}[!t]
\centering
\includegraphics[width=1\textwidth]{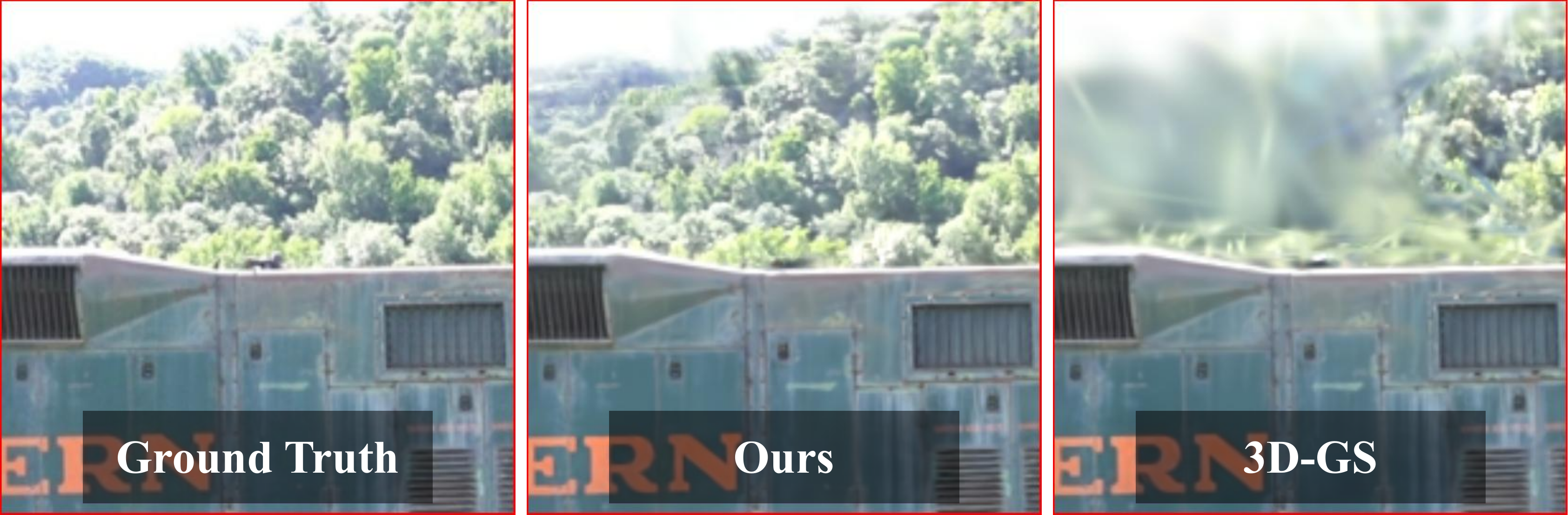}
\caption{
By minimizing the projection error through error analysis, we have achieved an improvement in the rendering image quality compared to the original 3D-GS~\cite{kerbl20233d}. These images are cropped from the complete images of the \textsc{Train} scene, in order to be highlighted.}
\label{fig:teaser}
% \vspace{-0.4cm}
\end{figure}

\section{Introduction}

Reconstructing 3D scenes from 2D images and synthesizing novel views has been a critical task in computer vision and graphics.
Recently, the trend of novel view synthesis is spearheaded by Neural Radiance Fields (NeRFs)~\cite{mildenhall2020nerf} and its variants~\cite{barron2021mip,barron2022mip}, which achieves photo-realistic rendering quality. NeRF represents scenes with Multi-Layer Perceptron (MLP) and renders novel views using volumetric ray-marching. 
However, these MLP-based methods contain a relatively large number of layers, leading to lengthy training and rendering times, which still fall significantly short of real-time rendering. 
To accelerate rendering, many methods resort to auxiliary data structures, such as hash grids~\cite{muller2022instant}, points or voxel~\cite{DBLP:conf/cvpr/TakikawaLYKLNJM21, fridovich2022plenoxels, DBLP:conf/cvpr/0004SC22, DBLP:conf/cvpr/XuXPBSSN22}. 
However, due to the dense sampling of points along rays, they still face challenges in achieving real-time performance.

With a substantial demand for high-speed and photo-realistic rendering effects, 3D Gaussian Splatting (3D-GS)~\cite{kerbl20233d} has emerged as an efficient representation that can be rendered at high speed on the GPU.
It has departed from the implicit scene representation using MLPs and instead opted for an explicit representation using Gaussian functions. As a result, 3D-GS avoids the expensive cost of densely sampling points in volume rendering, thus achieving real-time performance. 
Given a set of images with the camera parameters calibrated by SfM~\cite{schonberger2016structure}, 3D-GS aims to optimize a set of 3D Gaussians as graphical primitives to explicitly represent the scene. 
% \textit{Each 3D Gaussian is characterized by its position (mean) $\boldsymbol{\mu}$, covariance matrix $\boldsymbol{\Sigma}$, and opacity $\alpha$, and carries anisotropic spherical harmonics (SH) representing directional appearance component (color) of the radiance field, following standard practice \cite{muller2022instant, fridovich2022plenoxels}. 
% We initialize the set of 3D Gaussians with the sparse point cloud for free as part of the SfM process.}
To render 2D images, these 3D Gaussians are projected onto the image plane ($z=1$ plane) via a projection function denoted as $\phi$ for differentiable rasterization.

Unfortunately, Gaussian functions are closed under affine transformations but not under projective transformations. Therefore, 3D-GS adopts a local affine approximation~\cite{zwicker2002ewa}, specifically approximating the projection function with the first two terms of its Taylor expansion. 
Nevertheless, approximations introduce errors, and the local affine approximation similarly contributes to these errors which may lead to artifacts in the rendered images. 
This paper exploits the relationship between the error of 3D-GS and the Gaussian mean through the analysis of the Taylor remainder term. 
Additionally, it identifies the circumstances under which the error is minimized by determining the extremum of the error function.

Finally, based on the extremum analysis of the error function, we propose a novel optimal projection method to minimize the projection error in 3D-GS. This method requires only minor code modifications, does not affect real-time rendering performance, and yet achieves a significant improvement in rendering quality. Specifically, we project along the direction from the Gaussian mean to the camera center, where the projection plane is tangent to the line connecting the Gaussian mean and the camera center. 
Compared with the straightforward approach of projecting 3D Gaussians onto the image plane of the camera, the proposed projection reduces expected error while maintaining real-time and realistic rendering performance. 
Furthermore, we find that the projection errors in 3D-GS tend to escalate with the continuous expansion of the field of view and the corresponding reduction in focal length. This increasing error leads to more and larger elongated Gaussians, as well as cloud-like Gaussian artifacts, consequently degrading the overall image quality. In contrast, our method exhibits greater realism and robustness. As illustrated in Figure~\ref{fig:teaser}, our method significantly reduces blurriness and artifacts caused by the projection error.
Through experiments on various datasets and scenes, our method consistently outperforms the original projection method, particularly in settings with short focal lengths. And the method is easily implemented and adaptable to various camera models.
To summarize, we provide the following contributions: 
\begin{itemize}
    \item We provide thorough error analysis for 3D-GS that may lead to artifacts and consequently degrade rendering quality. We have identified the correlation between this error and the Gaussian position.
    \item We derive the mathematical expectation of this error function and analyze when this function takes extrema through methods of function optimization.
    \item We propose a novel optimal projection capable of adapting to diverse camera models that adopts different tangent plane projections based on the directions from each Gaussian mean to the camera center instead of naively projecting all Gaussians onto the same plane
\end{itemize}

\section{Related Work}

Novel view synthesis is a longstanding challenge in computer vision and graphics.  From traditional techniques ~\cite{schonberger2016structure, DBLP:conf/siggraph/GortlerGSC96, DBLP:conf/siggraph/LevoyH96, DBLP:conf/siggraph/BuehlerBMGC01,Sfm_init,DBLP:conf/iccv/GoeseleSCHS07, DBLP:journals/tog/ChaurasiaDSD13}, to neural network-based scene representations~\cite{jiang2020,deepsdf,genova2020,occupancynet,diffvolumetric,srn,DBLP:journals/tog/KopanasLRJD22}, various approaches have struggled to address the problem of synthesizing a new view from captured images. 

% \subsection{Traditional Scene Reconstruction and Rendering}

% Early novel-view synthesis methods were based on light fields  ~\cite{DBLP:conf/siggraph/GortlerGSC96, DBLP:conf/siggraph/LevoyH96}, transitioning from densely sampled to unstructured capture ~\cite{DBLP:conf/siggraph/BuehlerBMGC01}. Structure-from-Motion (SfM)~\cite{Sfm_init} introduced a new era using photo collections for view synthesis. Multi-view stereo (MVS) and subsequent algorithms ~\cite{DBLP:conf/iccv/GoeseleSCHS07, DBLP:journals/tog/ChaurasiaDSD13} blended input images but faced challenges. Recent neural rendering algorithms~\cite{DBLP:journals/tog/KopanasLRJD22} have overcome these issues, outperforming traditional methods.

\subsection{Neural Radiance Fields}

The Neural Radiance Field (NeRF)~\cite{mildenhall2020nerf} stands out as a successful neural rendering method based on MLPs, primarily owing to its encoding of position and direction. This encoding allows for effective reconstruction of high-frequency information in scenes. Notable improvements on this encoding have been made by MipNeRF~\cite{barron2021mip}, NeRF-W~\cite{nerfw}, FreeNeRF~\cite{freenerf}, and Instant NGP~\cite{muller2022instant}. These enhancements enable the handling of multi-resolution image inputs, multi-illumination with occlusion image inputs, sparse-view inputs, and achieve nearly real-time rendering capabilities, respectively. 
% Furthermore, NeRF++~\cite{nerfpp} analyzes the success factors of NeRF and introduces the concept of Foreground-Background NeRF. 
Barron et al. introduced MipNeRF360~\cite{barron2022mip} as an extension of MipNeRF~\cite{barron2021mip} to address the issue of generating low-quality renderings for unbounded scenes in NeRF. Importantly, these methods thoroughly exploit the intrinsic capabilities of NeRF as an implicit scene representation without introducing additional model priors. 

However, due to the implicit representation of scenes and the dense sampling of points along rays, they still face challenges in achieving real-time performance. 

\subsection{3D Gaussian Splatting}

3D Gaussian Splatting (3D-GS)~\cite{kerbl20233d} has demonstrated notable advancements in rendering performance by departing from MLPs and ray sampling, opting instead for Gaussian functions and Gaussian Splatting. This paradigm shift has attracted considerable attention within the industry, leading to various studies building upon this methodology~\cite{chen2024survey}.

3D-GS finds application in SLAM systems~\cite{gs_slam}, scene editing~\cite{gs_scene_edit} and segmentation~\cite{gs_scene_seg}. Efforts~\cite{gs_and_diffusion} have also been made to integrate 3D-GS with popular diffusion models~\cite{ddpm, diffusion2, diffusion3} and dynamic scenes~\cite{dynamic_gs}. Recent endeavors aim to enhance the robustness in sparse-view scenarios~\cite{xiong2023sparsegs,yang2024gaussianobject}, performance, storage efficiency~\cite{niedermayr2023compressed} and mesh reconstruction~\cite{sugar} of 3D-GS. However, these improvements do not specifically address errors associated with Gaussian projection. Potential methods for enhancing Gaussian functions include~\cite{mip_gs, radl2024stopthepop}. Nevertheless, they introduce a Gaussian filter and a novel Gaussian depth sorting algorithm, respectively, without investigating the projection function and associated errors. These approaches still rely on projecting onto the $z=1$ plane.

This paper aims to explore the potential of 3D Gaussian Splatting by analyzing errors that may arise during the projection process. The analysis will delve into the factors contributing to these errors and propose methods to mitigate and reduce them.

\section{Preliminaries}

\label{sec:preliminaries}

Similar to NeRF~\cite{mildenhall2020nerf}, the input to 3D-GS consists of a set of images, together
with the corresponding cameras calibrated by SFM~\cite{schonberger2016structure}. However, in contrast to NeRF, 3D-GS takes the sparse point cloud generated during the SFM calibration process as input. From these points, it  constructs a set of 3D Gaussians as graphical primitives explicitly representing the scene. Each 3D Gaussian is characterized by its position (mean) $\boldsymbol{\mu}$, covariance matrix $\boldsymbol{\Sigma}$, and opacity $\alpha$, and carries anisotropic spherical harmonics (SH) representing directional appearance component (color) of the radiance field, following
standard practice \cite{muller2022instant, fridovich2022plenoxels}. Subsequently, these 3D Gaussians are projected onto the image plane ($z=1$ plane) via a projection function for differentiable rasterization. Next, we will provide a detailed explanation of the Gaussian projection process in 3D-GS.

In the world space, we define a 3D Gaussian function by $G$, characterized by its mean $\boldsymbol{\mu}$ and covariance matrix $\boldsymbol{\Sigma}$.
% expressed as:
% \begin{equation}
% G_{\mathbf{\mu}, \boldsymbol{\Sigma}}\left(\mathbf{x}\right)=\exp\left(-1/2\left(\mathbf{x}-\boldsymbol{\mu}\right)^\top\boldsymbol{\Sigma}^{-1}\left(\mathbf{x}-\boldsymbol{\mu}\right)\right)\text{.}
% \label{eq:gaussian}
% \end{equation}
To project the Gaussian onto the image plane, the initial step in 3D-GS involves an affine transformation of this Gaussian function from the world space to the camera space via a viewing transformation matrix $\mathbf{W}$, yielding a transformed Gaussian function:
\begin{equation}
\begin{split}
G\left(\mathbf{x}\right)=\exp\left\{-1/2\left(\mathbf{W}\mathbf{x}-\mathbf{W}\boldsymbol{\mu}\right)^\top
\left(\mathbf{W}\boldsymbol{\Sigma}\mathbf{W}^{\top}\right)^{-1}
\left(\mathbf{W}\mathbf{x}-\mathbf{W}\boldsymbol{\mu}\right)\right\}\text{.}
\end{split}
\label{eq:view_transformation}
\end{equation} 
% which can also be represented by the following function:
% \begin{equation}
% \begin{split}
% G^{'}\left(\mathbf{x^{'}}\right)&=\exp\left(-1/2\left(\mathbf{x^{'}}-\boldsymbol{\mu^{'}}\right)^\top
% {\boldsymbol{\Sigma}^{'}}^{-1}
% \left(\mathbf{x^{'}}-\boldsymbol{\mu^{'}}\right)\right)
% \end{split}
% \label{eq:view_transformation_2}
% \end{equation} where 
We denote $G^{'}$ as the 3D Gaussian in the camera space, with $\mathbf{x}^{'}=\mathbf{W}\mathbf{x}$, $\boldsymbol{\mu}^{'}=\mathbf{W}\boldsymbol{\mu}$ and $\boldsymbol{\Sigma}^{'}=\mathbf{W}\boldsymbol{\Sigma}\mathbf{W}^{\top}$ as the point, mean and covariance matrix in the camera space, respectively.

Subsequently, traditional 3D-GS requires projecting the Gaussian function in the camera space onto the $z=1$ plane, 
% The projection plane is formulated as:
% \begin{equation}
%     \mathbf{x_0}^\top\cdot\left(\mathbf{x}^{'}-\mathbf{x_0}\right)=0
% \label{eq:zplane_eq}
% \end{equation} 
with the projection function $\varphi$ defined as:
\begin{equation}
    \varphi\left(\mathbf{x}^{'}\right)=\mathbf{x}^{'} \left(\mathbf{x_0}^{\top} \mathbf{x}^{'}\right)^{-1} \left(\mathbf{x_0}^{\top} \mathbf{x_0}\right)=\mathbf{x}^{'} \left(\mathbf{x_0}^{\top} \mathbf{x}^{'}\right)^{-1}
\end{equation} where $\mathbf{x_0}=\left[\begin{matrix}
    0, 0, 1
\end{matrix}\right]^{\top}$ represents the projection of the camera space's origin onto this plane. Expanding this projection function to the first order using Taylor series yields:
\begin{equation}
\begin{split}
\varphi\left(\mathbf{x}^{'}\right)&=\varphi\left(\boldsymbol{\mu}^{'}\right)+\frac{\partial \varphi}{\partial \mathbf{x}^{'}}\left(\boldsymbol{\mu}^{'}\right)\left(\mathbf{x}^{'}-\boldsymbol{\mu}^{'}\right)+R_1\left(\mathbf{x}^{'}\right)\\
&\approx\varphi\left(\boldsymbol{\mu}^{'}\right)+\frac{\partial \varphi}{\partial \mathbf{x}^{'}}\left(\boldsymbol{\mu}^{'}\right)\left(\mathbf{x}^{'}-\boldsymbol{\mu}^{'}\right) 
\end{split}
\label{eq:Taylor}
\end{equation} where $\frac{\partial \varphi}{\partial \mathbf{x}^{'}}\left(\boldsymbol{\mu}^{'}\right)$  is the Jacobian of the affine approximation of the projective transformation, denoted as $\mathbf{J}$.
Applying this local affine approximation by neglecting the Taylor remainder term allows to derive the 2D Gaussian function $G_{\text{2D}}$ projected onto the $z=1$ plane as:
\begin{equation}
\begin{split}
G_{\text{2D}}\left(\mathbf{x}^{'}\right)&=\exp\left\{-1/2\left(\mathbf{J}\mathbf{x}^{'}-\mathbf{J}\boldsymbol{\mu}^{'}\right)^\top
\left(\mathbf{J}\boldsymbol{\Sigma}^{'}\mathbf{J}^{\top}\right)^{-1}
\left(\mathbf{J}\mathbf{x}^{'}-\mathbf{J}\boldsymbol{\mu}^{'}\right)\right\}\\
&\approx\exp\left\{-1/2\left(\varphi\left(\mathbf{x}^{'}\right)-\varphi\left(\boldsymbol{\mu}^{'}\right)\right)^\top
\left(\mathbf{J}\boldsymbol{\Sigma}^{'}\mathbf{J}^{\top}\right)^{-1}
\left(\varphi\left(\mathbf{x}^{'}\right)-\varphi\left(\boldsymbol{\mu}^{'}\right)\right)\right\}.
\end{split} 
\label{eq:proj_transformation}
\end{equation} 

Since the rank of the matrix $\mathbf{J}$ is 2, the inverse of $\mathbf{J}\boldsymbol{\Sigma}^{'}\mathbf{J}^{\top}$
  is, in fact, the inverse of the covariance matrix of the 2D Gaussian obtained by skipping the third row and column~\cite{zwicker2002ewa}. 
%   Similarly, this function can be expressed in another form:
% \begin{equation}
% G_{\text{2D}}\left(\mathbf{x_{\text{2D}}}\right)=\exp\left(-1/2\left(\mathbf{x_{\text{2D}}}-\boldsymbol{\mu_{\text{2D}}}\right)^\top
% \boldsymbol{\Sigma}_{\text{2D}}^{-1}
% \left(\mathbf{x_{\text{2D}}}-\boldsymbol{\mu_{\text{2D}}}\right)\right)
% \end{equation} where 
We denote $\mathbf{x}_{\text{2D}}=\varphi\left(\mathbf{x}^{'}\right)$, $\boldsymbol{\mu}_{\text{2D}}=\varphi\left(\boldsymbol{\mu}^{'}\right)$ and $\boldsymbol{\Sigma}_{\text{2D}}=\mathbf{J}\boldsymbol{\Sigma}^{'}\mathbf{J}^{\top}$ as the point, mean and $2\times2$ covariance matrix of $G_{\text{2D}}$ in the image space, respectively. For a pixel $(u, v)$, we obtain the value of the 2D Gaussian function by querying $\mathbf{x_{\text{2D}}} = \left[\begin{matrix}u, v
\end{matrix}\right]^{\top}$, and then perform alpha blending to derive the color of the pixel. The blue box in Figure~\ref{fig:optimal_proj} illustrates the projection process.

% After the aforementioned steps, 3D-GS projects the 3D Gaussian from the world coordinate system onto the image plane. Subsequently, rasterization is performed on the image plane to obtain the rendered image. 

\section{Local Affine Approximation Error}

\label{sec:error}

From Equation \ref{eq:Taylor} and Equation \ref{eq:proj_transformation} we observe that 3D Gaussian Splatting introduces an approximation during the projection transformation, i.e. the local affine approximation. Consequently, 3D-GS actually adopts a 2D Gaussian function that is not the true projection of the initial 3D Gaussian. The error introduced by this approximation can be characterized by the Taylor remainder term in Equation \ref{eq:Taylor}:
\begin{equation}
\begin{split}
&R_1\left(\mathbf{x}^{'}\right)=\varphi\left(\mathbf{x}^{'}\right)-\varphi\left(\boldsymbol{\mu}^{'}\right)-\frac{\partial \varphi}{\partial \mathbf{x}^{'}}\left(\boldsymbol{\mu}^{'}\right)\left(\mathbf{x}^{'}-\boldsymbol{\mu}^{'}\right)
\label{eq:error_vec}
\end{split}
\end{equation} where 
\begin{equation}
\frac{\partial \varphi}{\partial \mathbf{x}^{'}}\left(\boldsymbol{\mu}^{'}\right)=\mathbb{I} \otimes \left(\mathbf{x_0}^{\top} \boldsymbol{\mu}^{'}\right)^{-1}  -  {\boldsymbol{\mu}^{'}}\left(\mathbf{x_0}^{\top} \boldsymbol{\mu}^{'}\right)^{-1} \left({\boldsymbol{\mu}^{'}}^{\top} \mathbf{x_0}\right)^{-1} \mathbf{x_0}^{\top}\text{.}
\label{eq:J}
\end{equation}$\mathbb{I}$ represents the identity matrix and $\otimes$ denotes the multiplication of a matrix and a scalar. Clearly, this Taylor remainder term is a three-dimensional vector related to the random variable $\mathbf{x}^{'}$ and the mean $\boldsymbol{\mu}^{'}$ of the Gaussian function. Therefore, by computing the square of the Frobenius norm of this vector $\Vert R_1\left(\mathbf{x}^{'}\right)\Vert_{F}^2$ and taking the mathematical expectation of this norm function with respect to the random variable $\mathbf{x}^{'}$, we obtain an error function that depends solely on the Gaussian mean $\boldsymbol{\mu}^{'}$:
\begin{equation}
\epsilon\left(\boldsymbol{\mu}^{'}\right)=\int\limits_{\mathbf{x}^{'}\in\mathcal{X}^{'}} \Vert R_1\left(\mathbf{x}^{'}\right)\Vert_{F}^2 \mathrm{d}\mathbf{x}^{'}\text{.}
\label{eq:error_function}
\end{equation}

% Before seeking the extremum of the error function, let's simplify the integral expression. First, we simplify $\mathbf{x}^{'}$ and $\boldsymbol{\mu}^{'}$ to unit vectors projected onto the unit sphere centered at the camera center (hereafter referred to as the unit sphere) by proving that the composition of transformations $\left(\phi\circ\pi\right)$, projecting a point in three-dimensional space onto the unit sphere $\pi$ and subsequently onto the projection plane $\phi$, is equivalent to directly projecting a point in three-dimensional space onto the projection plane. This simplification ensures that the entire error function involves only operations between three unit space vectors. The proof process for this is presented in the Supplementary Materials.
% \begin{equation}
% \begin{split}
% \left(\phi\circ\pi\right)\left({\mathbf{x}^{'}}\right)&=\phi\left({\mathbf{x}^{'}}{\left({\mathbf{x}^{'}}^{\top}{\mathbf{x}^{'}}\right)^{-1/2}}\right)\\
% &={\mathbf{x}^{'}}{\left({\mathbf{x}^{'}}^{\top}{\mathbf{x}^{'}}\right)^{-1/2}}\left(\mathbf{x_0}^{\top}\left({\mathbf{x}^{'}}{\left({\mathbf{x}^{'}}^{\top}{\mathbf{x}^{'}}\right)^{-1/2}}\right)\right)^{-1}\\
% &={\mathbf{x}^{'}} \left(\mathbf{x_0}^{T}{\mathbf{x}^{'}}\right)^{-1}\\
% &=\phi\left({\mathbf{x}^{'}}\right)\text{.}
% \end{split}
% \end{equation}

The simplified error function $\epsilon$ involves three unit vectors: $\mathbf{x}^{'}$, $\mathbf{x_0}$ and $\boldsymbol{\mu}^{'}$.
% , as illustrated in Figure \ref{fig:overview_phi}. 
Their spherical coordinates are given as follows:
\begin{equation}
\mathbf{x_0}=\left[\begin{matrix}\sin{\left(\phi_0 \right)} \cos{\left(\theta_0 \right)}\\- \sin{\left(\theta_0 \right)}\\\cos{\left(\phi_0 \right)} \cos{\left(\theta_0 \right)}\end{matrix}\right],
\mathbf{x}^{'}=\left[\begin{matrix}\sin{\left(\phi \right)} \cos{\left(\theta \right)}\\- \sin{\left(\theta \right)}\\\cos{\left(\phi \right)} \cos{\left(\theta \right)}\end{matrix}\right],
\boldsymbol{\mu}^{'}=\left[\begin{matrix}\sin{\left(\phi_{\mu} \right)} \cos{\left(\theta_{\mu} \right)}\\- \sin{\left(\theta_{\mu} \right)}\\\cos{\left(\phi_{\mu} \right)} \cos{\left(\theta_{\mu} \right)}\end{matrix}\right]
\label{eq:x_0_x_mu_cos}
\end{equation} where $\phi_0=0$ and $\theta_0=0$. Please refer to the detailed proofs in the supplementary materials. 
% \begin{equation}
% \mathbf{J}=\frac{\partial \phi}{\partial \mathbf{x}^{'}}\left(\boldsymbol{\mu}^{'}\right)=
% \left[\begin{matrix}\frac{1}{\cos{\left(\phi_{\mu} \right)} \cos{\left(\theta_{\mu} \right)}} & 0 & - \frac{\sin{\left(\phi_{\mu} \right)}}{\cos^{2}{\left(\phi_{\mu} \right)} \cos{\left(\theta_{\mu} \right)}}\\0 & \frac{1}{\cos{\left(\phi_{\mu} \right)} \cos{\left(\theta_{\mu} \right)}} & \frac{\sin{\left(\theta_{\mu} \right)}}{\cos^{2}{\left(\phi_{\mu} \right)} \cos^{2}{\left(\theta_{\mu} \right)}}\\0 & 0 & 0\end{matrix}\right]
% \end{equation}
% \begin{equation}
% \label{eq:error_vec_sphere}
% R_1\left(\mathbf{x}^{'}\right)
% =\left[\begin{matrix}- \frac{\sin{\left(\phi - \phi_{\mu} \right)} \cos{\left(\theta \right)}}{\cos^{2}{\left(\phi_{\mu} \right)} \cos{\left(\theta_{\mu} \right)}} + \tan{\left(\phi \right)} - \tan{\left(\phi_{\mu} \right)}\\\frac{\sin{\left(\theta \right)}}{\cos{\left(\phi_{\mu} \right)} \cos{\left(\theta_{\mu} \right)}} - \frac{\sin{\left(\theta_{\mu} \right)} \cos{\left(\phi \right)} \cos{\left(\theta \right)}}{\cos^{2}{\left(\phi_{\mu} \right)} \cos^{2}{\left(\theta_{\mu} \right)}} + \frac{\tan{\left(\theta_{\mu} \right)}}{\cos{\left(\phi_{\mu} \right)}} - \frac{\tan{\left(\theta \right)}}{\cos{\left(\phi \right)}}\\0\end{matrix}\right]\text{.}
% \end{equation}

\begin{figure}[!t]    
  \centering          
  % \subfloat[$\left[-\pi/4\cdot0.0095, \pi/4\cdot0.0095\right]$]  
  % {
  %     \label{fig:epsilon_0095}\includegraphics[width=0.33\textwidth]{figures/overview/epsilon_0095.pdf}
  % }
  \subfloat [$\lambda=0.095$] {
  \label{fig:epsilon_095}
\includegraphics[width=0.35\textwidth]{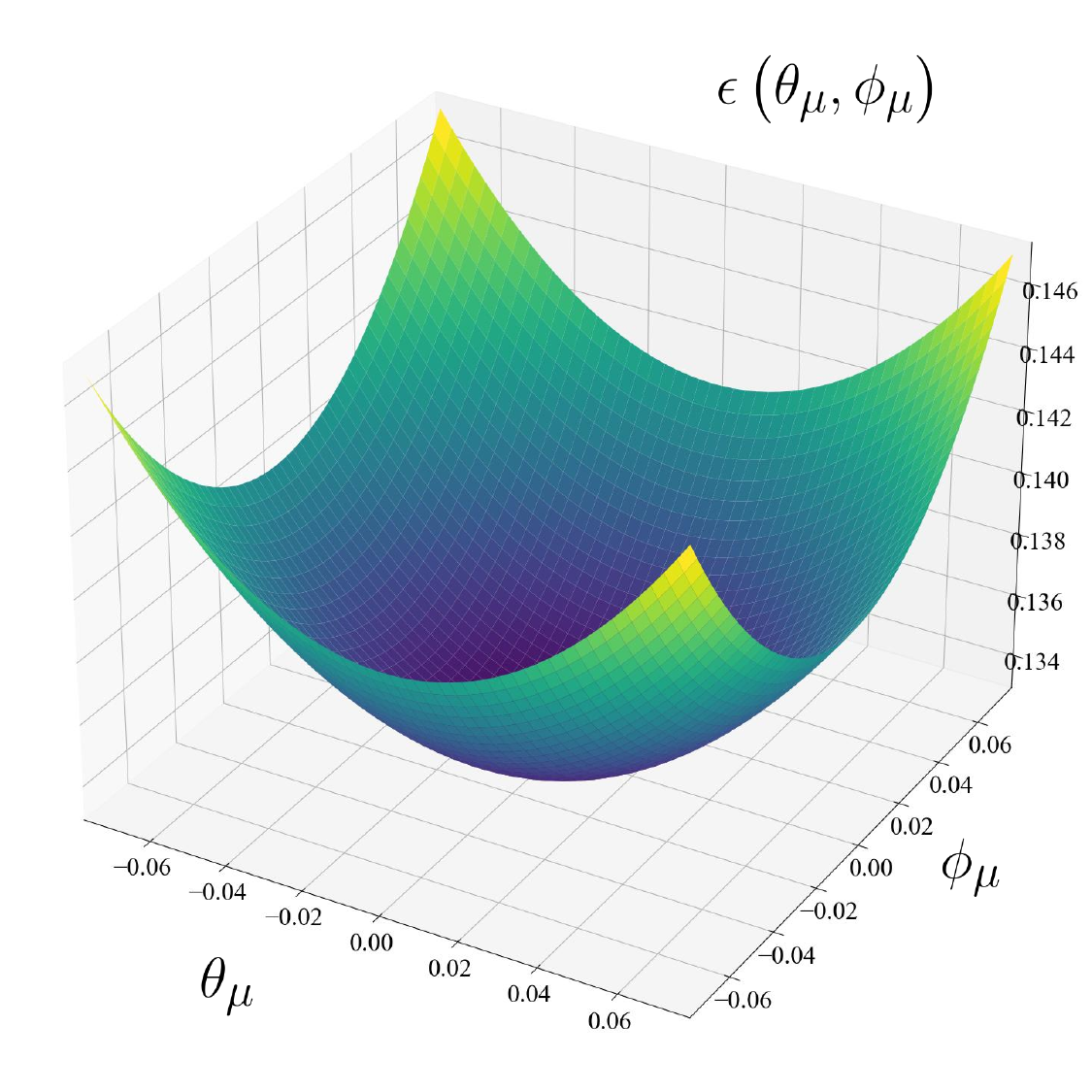}
  }\hspace{0.8cm}
  \subfloat [$\lambda=0.95$] {
  \label{fig:epsilon_95}
\includegraphics[width=0.35\textwidth]{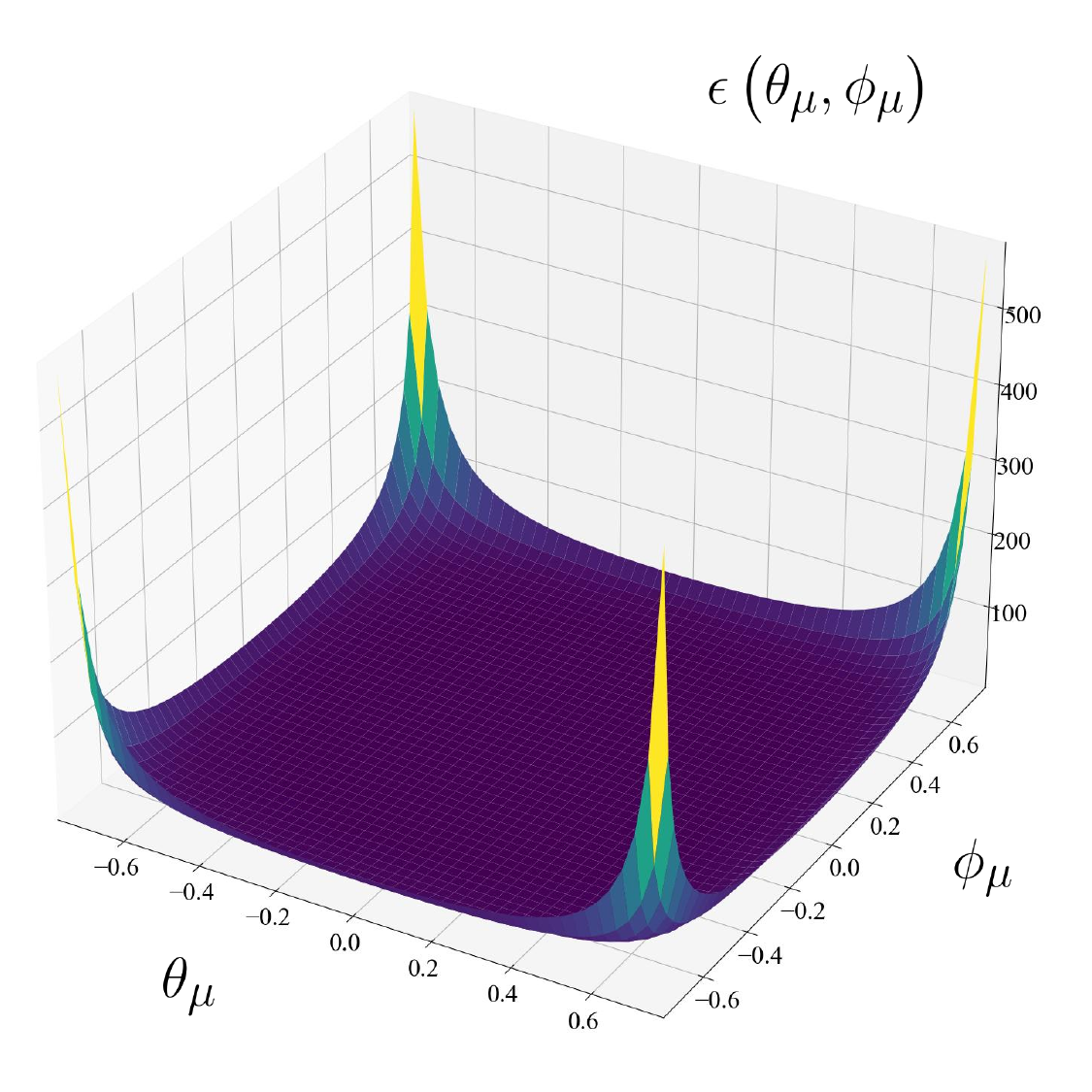}
  }
  \caption{
Visualization of the 3D Gaussian Splatting error function $\epsilon\left(\theta_{\mu},\phi_{\mu}\right)$ under two distinct domains. 
The domain of the function is $\left\{\left(\theta_{\mu},\phi_{\mu}\right)\mid
\theta_{\mu}\in\left[-\lambda\pi/4, \lambda\pi/4\right]
 ~\land~\phi_{\mu}\in\left[-\lambda\pi/4, \lambda\pi/4\right]
\right\}$ where $\left(\theta_{\mu},\phi_{\mu}\right)$ represents the polar coordinate of the 3D Gaussian mean and $\lambda$ represents the scaling factor of the domain.}   
  \label{fig:epsilon}   
  % \vspace{-0.4cm}
\end{figure}

The size of the integration region is related to the covariance of the Gaussian function. Since the current discussion specifically focuses on the impact of the mean on the error rather than the variance, let's assume the integration region to be $\left\{\mathbf{x}^{'}\mid
\theta\in\left[-\pi/4+\theta_{\mu}, \pi/4+\theta_{\mu}\right]
 ~\land~\phi\in\left[-\pi/4+\phi_{\mu}, \pi/4+\phi_{\mu}\right]
\right\}$. Then, based on Equation \ref{eq:error_vec}-\ref{eq:x_0_x_mu_cos}, we obtain:
\begin{equation}
\begin{split}
\epsilon\left(\theta_{\mu}, \phi_{\mu}\right)&=\int_{-\pi/4+\theta_{\mu}}^{\pi/4+\theta_{\mu}}\int_{-\pi/4+\phi_{\mu}}^{\pi/4+\phi_{\mu}} 
\left(- \frac{\sin{\left(\phi - \phi_{\mu} \right)} \cos{\left(\theta \right)}}{\cos^{2}{\left(\phi_{\mu} \right)} \cos{\left(\theta_{\mu} \right)}} + \tan{\left(\phi \right)} - \tan{\left(\phi_{\mu} \right)}\right)^2
+\\
&\left(
\frac{\sin{\left(\theta \right)}}{\cos{\left(\phi_{\mu} \right)} \cos{\left(\theta_{\mu} \right)}} - \frac{\sin{\left(\theta_{\mu} \right)} \cos{\left(\phi \right)} \cos{\left(\theta \right)}}{\cos^{2}{\left(\phi_{\mu} \right)} \cos^{2}{\left(\theta_{\mu} \right)}} + \frac{\tan{\left(\theta_{\mu} \right)}}{\cos{\left(\phi_{\mu} \right)}} - \frac{\tan{\left(\theta \right)}}{\cos{\left(\phi \right)}}
\right)^2
\mathrm{d}\theta\mathrm{d}\phi\text{.}
\end{split}
\label{eq:error_function_thetaphi}
\end{equation} Fortunately, this integral has a close-form expression, but its form is very complicated. We report the close-form expression in the supplemental materials and visually depict it in Figure~\ref{fig:epsilon}.

% \vspace{-0.8cm}

The error function is partially differentiated with respect to $\theta_{\mu}$ and $\phi_{\mu}$. It is easy to check that at $\theta_{\mu}=\theta_0=0$ and $\phi_{\mu}=\phi_0=0$, the following holds:
\begin{equation}
\frac{\partial \epsilon }{\partial \theta_{\mu}}\left(0, 0\right)=0,
\frac{\partial \epsilon }{\partial \phi_{\mu}}\left(0, 0\right)=0\text{.}
\end{equation} Therefore, this point is actually an extremum point of the function. We also confirm that this point is the minimum of the function. Indeed, from Figure \ref{fig:epsilon_095}, it is also apparent that the function is concave, with its minimum value occurring at $(0,0)$. Additionally, it can be observed that the minimum value of the error function is greater than zero, due to the fact that Gaussian functions are not closed under projection transformations. Figure \ref{fig:epsilon_95} shows that in the majority of the region close to the origin, the function values vary very gently. However, as the independent variable of the function approaches the boundaries, the function values increase rapidly, resulting in a substantial difference between the maximum and minimum values. 

The above analysis indicates that the error is small and does not significantly affect the quality of rendered images. Therefore, traditional 3D-GS~\cite{kerbl20233d} utilizes local affine approximation but still successfully reconstructs the scene to obtain a high-quality novel view image. Nevertheless, the naive projection of all Gaussians onto the same plane $z=1$ in 3D-GS may lead to larger projection errors for Gaussians farther from the plane center (the projection point of the camera center to the plane), causing artifacts. These artifacts are severe when using wide-angle lenses, significantly degrading the quality of the renderings. 

\section{Optimal Gaussian Splatting}

% \vspace{-0.8cm}
\begin{figure}[!t]
\centering
\includegraphics[width=1\textwidth]{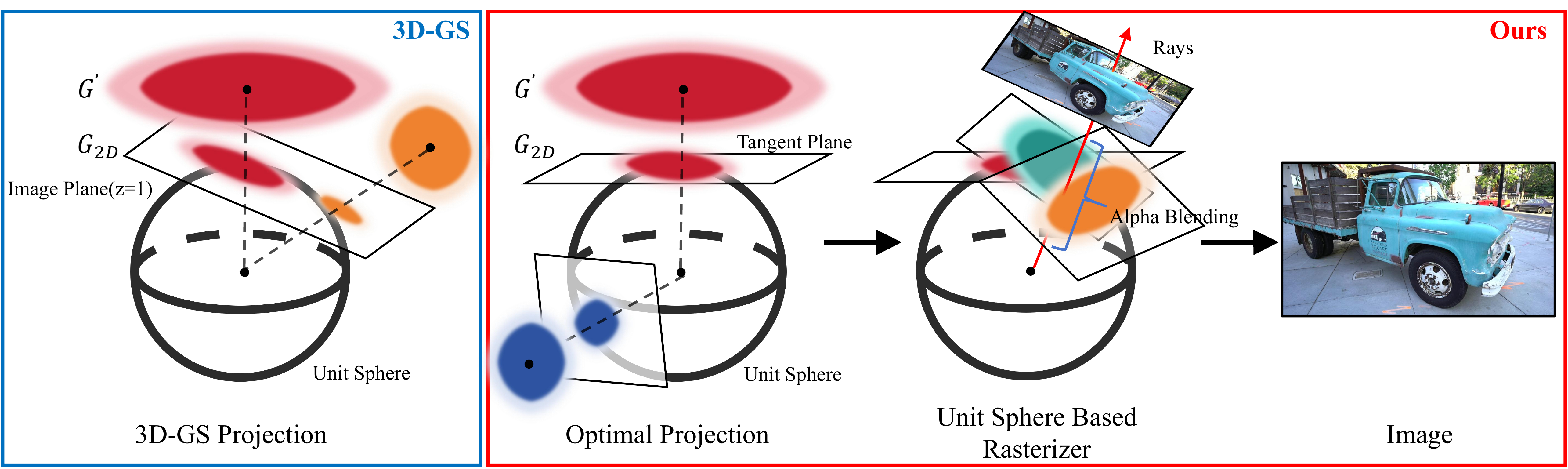}
\caption{Illustration of the rendering pipeline for our \textbf{Optimal Gaussian Splatting} and the projection of 3D-GS~\cite{kerbl20233d}. The blue box depicts the projection process of the original 3D-GS, which straightforwardly projects all Gaussians onto the same projection plane. In contrast, the red box illustrates our approach, where we project individual Gaussians onto corresponding tangent planes.}
\label{fig:optimal_proj}
% \vspace{-0.4cm}
\end{figure}

% The first step, similar to the original 3D-GS~\cite{kerbl20233d}, involves transforming world coordinates into the camera coordinate system. Instead of projecting Gaussians onto the z=1 plane, each Gaussian is radially projected based on its mean along the line connecting it to the camera center, projecting onto a plane tangent to the unit sphere and perpendicular to this line. After projection, colors of points on the unit sphere are computed by alpha blending the 2D Gaussians on the tangent plane. Finally, for each pixel in the image, rays are cast onto the unit sphere  to retrieve the color of that pixel, resulting in the rendered image.

Based on the analysis of the error function, we introduce Optimal Gaussian Splatting shown in Figure \ref{fig:optimal_proj}. This helps us to achieve smaller projection errors, resulting in higher-quality rendering. The first step, similar to the original 3D-GS~\cite{kerbl20233d}, involves transforming world coordinates into the camera coordinate system. Instead of projecting Gaussians onto the $z=1$ plane, each Gaussian is radially projected based on its mean along the line connecting it to the camera center, projecting onto a plane tangent to the unit sphere and perpendicular to this line. After projection, colors of points on the unit sphere are computed by alpha blending the 2D Gaussians on the tangent plane. Finally, for each pixel in the image, rays are cast onto the unit sphere  to retrieve the color of that pixel, resulting in the rendered image.

\subsection{Optimal Projection}

We discover that the error function attains its minimum value when the projection of the Gaussian mean on the plane coincides with the projection from the camera center to the plane. Therefore, in Optimal Gaussian Splatting, we employ an optimal projection method. Specifically, instead of naively projecting different Gaussians onto the same plane $z=1$, we adopt distinct projection planes for each Gaussian. These projection planes are determined by the tangent planes formed by the Gaussian mean and the line connecting it to the camera center. The specific projection plane is formulated as:
\begin{equation}
\mathbf{x_{p}}^\top\cdot\left(\mathbf{x}^{'}-\mathbf{x_{p}}\right)=0
        \label{eq:tangentplane_eq}
\end{equation}
where
\begin{equation}
\mathbf{x_{p}}=\varpi\left({\boldsymbol{\mu}^{'}}\right)={\boldsymbol{\mu}^{'}}\left({\boldsymbol{\mu}^{'}}^{\top}{\boldsymbol{\mu}^{'}}\right)^{-1/2}
\end{equation} represents the projection of the camera space's origin onto this plane and $\varpi$ function represents the transformation that projects points onto the unit sphere. According to Equation \ref{eq:tangentplane_eq}, the optimal projection function $\varphi_{\text{p}}$ is obtained as:
\begin{equation}
    \varphi_{\text{p}}\left(\mathbf{x}^{'}\right)=\mathbf{x}^{'} \left(\mathbf{x_{p}}^{\top} \mathbf{x}^{'}\right)^{-1} \left(\mathbf{x_{p}}^{\top} \mathbf{x_{p}}\right)=\mathbf{x}^{'} \left(\mathbf{x_{p}}^{\top} \mathbf{x}^{'}\right)^{-1}\text{.}
\end{equation} The corresponding local affine approximation Jacobian matrix $\mathbf{J_{p}}$ is:
\begin{equation}
\mathbf{J_{p}}=\frac{\partial \varphi_{\text{p}}}{\partial \mathbf{x}^{'}}\left(\boldsymbol{\mu}^{'}\right)=\mathbb{I} \otimes \left(\mathbf{x_{p}}^{\top} \boldsymbol{\mu}^{'}\right)^{-1}  -  {\boldsymbol{\mu}^{'}}\left(\mathbf{x_{p}}^{\top} \boldsymbol{\mu}^{'}\right)^{-1} \left({\boldsymbol{\mu}^{'}}^{\top} \mathbf{x_{p}}\right)^{-1} \mathbf{x_{p}}^{\top}\text{.}
\label{eq:J_mu}
\end{equation}
When $\boldsymbol{\mu}^{'}=\left[\begin{matrix}
    \mu_x,\mu_y,\mu_z
\end{matrix}\right]^{\top}$, the specific form of this Jacobian matrix is:
\begin{equation}
\mathbf{J_{p}}=\left[\begin{matrix}\frac{\mu_{y}^{2} + \mu_{z}^{2}}{\left(\mu_{x}^{2} + \mu_{y}^{2} + \mu_{z}^{2}\right)^{\frac{3}{2}}} & - \frac{\mu_{x} \mu_{y}}{\left(\mu_{x}^{2} + \mu_{y}^{2} + \mu_{z}^{2}\right)^{\frac{3}{2}}} & - \frac{\mu_{x} \mu_{z}}{\left(\mu_{x}^{2} + \mu_{y}^{2} + \mu_{z}^{2}\right)^{\frac{3}{2}}}\\- \frac{\mu_{x} \mu_{y}}{\left(\mu_{x}^{2} + \mu_{y}^{2} + \mu_{z}^{2}\right)^{\frac{3}{2}}} & \frac{\mu_{x}^{2} + \mu_{z}^{2}}{\left(\mu_{x}^{2} + \mu_{y}^{2} + \mu_{z}^{2}\right)^{\frac{3}{2}}} & - \frac{\mu_{y} \mu_{z}}{\left(\mu_{x}^{2} + \mu_{y}^{2} + \mu_{z}^{2}\right)^{\frac{3}{2}}}\\- \frac{\mu_{x} \mu_{z}}{\left(\mu_{x}^{2} + \mu_{y}^{2} + \mu_{z}^{2}\right)^{\frac{3}{2}}} & - \frac{\mu_{y} \mu_{z}}{\left(\mu_{x}^{2} + \mu_{y}^{2} + \mu_{z}^{2}\right)^{\frac{3}{2}}} & \frac{\mu_{x}^{2} + \mu_{y}^{2}}{\left(\mu_{x}^{2} + \mu_{y}^{2} + \mu_{z}^{2}\right)^{\frac{3}{2}}}\end{matrix}\right]\text{.}
\end{equation} It's very easy to implement modifications to the Jacobian matrix in the forward process.

\subsection{Unit Sphere Based Rasterizer}
\label{subsec:rasterizer}

For the proposed Optimal Projection, we have proposed a Unit Sphere Based Rasterizer to generate images. Through Optimal Projection, we obtain the projection of the three-dimensional Gaussians on the tangent plane of the unit sphere instead of obtaining the Gaussians in the image space. Therefore, we need to rasterize based on this unit sphere. 

Specifically, for a pixel $\left(u, v\right)$ on the image, similar to NeRF, we cast a ray. However, unlike NeRF, we do not involve the extensive sampling of points along the ray, which would significantly degrade performance. Instead, our focus is solely on determining which tangent-plane Gaussians the ray intersects on the unit sphere:
\begin{equation}
\mathbf{x}_{\text{2D}}=\varphi_{\text{p}}\left(\left[\begin{matrix}
    \left(u - c_x\right)/f_x\\
    \left(v - c_y\right)/f_y\\
    1
    \end{matrix}\right]\right)
\label{eq:img2tan}
\end{equation} where $c_x$, $c_y$, $f_x$, $f_y$ denote the intrinsic parameters of the pinhole camera. We then query the function values of these Gaussians for alpha blending to obtain the color
\begin{equation}
G_{\text{2D}}\left(\mathbf{x_{\text{2D}}}\right)=\exp\left\{-1/2\left(\mathbf{x_{\text{2D}}}-\varphi_{\text{p}}\left(\boldsymbol{\mu}^{'}\right)\right)^\top
\left(
\mathbf{J_{p}}
\boldsymbol{\Sigma}^{'}
\mathbf{J_{p}}^{\top}
\right)^{-1}
\left(\mathbf{x_{\text{2D}}}-\varphi_{\text{p}}\left(\boldsymbol{\mu}^{'}\right)\right)\right\}\text{.}
\end{equation}
% \begin{equation}
% \alpha\left(u, v, \boldsymbol{\mu}\right)=\alpha_{\boldsymbol{\mu}}\cdot G^{\mu}_{\text{2D}}\left(\mathbf{x^{\mu}_{\text{2D}}}\right)
% \end{equation}
% \begin{equation}  \mathbf{C}\left(u,v\right)=\sum_{z=z_{\text{near}}}^{z_{\text{far}}}{
% \text{SH}\left(\boldsymbol{\mu}_z\right)\alpha\left(u, v, \boldsymbol{\mu}_z\right)\prod_{k=z_{\text{near}}}^{z}\left(1-\alpha\left(u, v, \boldsymbol{\mu}_{k}\right)\right)
% }
% \end{equation} where the summation and product from $z_\text{near}$ to $z_\text{far}$  represent discretized depths sorted in ascending order from near to far for the depths of 3D Gaussians encountered by the ray,  $\alpha_{\boldsymbol{\mu}}$ represents the transparency of the 3D Gaussian with its mean $\boldsymbol{\mu}$, $\boldsymbol{\mu}_z$ represents the mean of the Gaussian at depth $z$, and SH represents the spherical harmonics of this Gaussian.

Note that both $\mathbf{J_{p}}$ in Equation \ref{eq:J_mu} and $\mathbf{J}$ in Equation \ref{eq:J} are matrices with a rank of 2. Consequently, the inverse of the covariance matrix of the 2D Gaussian is also obtained by skipping the third row and column. However, the third row and third column of $\mathbf{J_{p}}
\boldsymbol{\Sigma}^{'}
\mathbf{J_{p}}^{\top}$ are not entirely zeros, as opposed to $\mathbf{J}
\boldsymbol{\Sigma}^{'}
\mathbf{J}^{\top}$. To ensure the equation still holds, an invertible matrix $\mathbf{Q}$
needs to be used for left multiplication with $\mathbf{x_{\text{2D}}}$, $\varphi_{\text{p}}\left(\boldsymbol{\mu}^{'}\right)$ and $\mathbf{J_p}$:
\begin{equation}
\mathbf{Q}=\left[\begin{matrix}\frac{\mu_{z}}{\sqrt{\mu_{x}^{2} + \mu_{z}^{2}}} & 0 & - \frac{\mu_{x}}{\sqrt{\mu_{x}^{2} + \mu_{z}^{2}}}\\- \frac{\mu_{x} \mu_{y}}{\sqrt{\mu_{x}^{2} + \mu_{z}^{2}} \sqrt{\mu_{x}^{2} + \mu_{y}^{2} + \mu_{z}^{2}}} & \frac{\sqrt{\mu_{x}^{2} + \mu_{z}^{2}}}{\sqrt{\mu_{x}^{2} + \mu_{y}^{2} + \mu_{z}^{2}}} & - \frac{\mu_{y} \mu_{z}}{\sqrt{\mu_{x}^{2} + \mu_{z}^{2}} \sqrt{\mu_{x}^{2} + \mu_{y}^{2} + \mu_{z}^{2}}}\\\frac{\mu_{x}}{\sqrt{\mu_{x}^{2} + \mu_{y}^{2} + \mu_{z}^{2}}} & \frac{\mu_{y}}{\sqrt{\mu_{x}^{2} + \mu_{y}^{2} + \mu_{z}^{2}}} & \frac{\mu_{z}}{\sqrt{\mu_{x}^{2} + \mu_{y}^{2} + \mu_{z}^{2}}}\end{matrix}\right]\text{.}
\end{equation}

Since our projection does not rely on the perspective image plane $z=1$ as in the original 3D-GS~\cite{kerbl20233d}, we can adapt to different camera models such as fisheye cameras and panoramic images by simply modifying the transformation from image space to camera space, as described in Equation~\ref{eq:img2tan}.

\section{Experiments}

\label{sec:experiments}

To validate the effectiveness of the Optimal Gaussian Splatting derived from theoretical error analysis, we conduct a series of experiments, comparing it with the original 3D-GS~\cite{kerbl20233d} and some current state-of-the-art methods.

\subsection{Implementation}

We implemented Optimal Gaussian Splatting based on the PyTorch framework in 3D-GS~\cite{kerbl20233d}  and wrote custom CUDA kernels for rasterization. We used the default parameters of 3D-GS to maintain consistency with the original 3D-GS and prevent other factors from introducing interference into the results.

\subsection{Datasets}

We test our algorithm on a total of 13 real scenes which are the same as those used in the original 3D-GS~\cite{kerbl20233d}. In particular, we evaluate our approach on the complete set of scenes featured in Mip-NeRF360~\cite{barron2022mip}, which currently represents the state-of-the-art in NeRF rendering quality. Additionally, we test our method on two scenes from the Tanks \& Temples dataset~\cite{knapitsch2017tanks} and two scenes provided by Hedman et al. ~\cite{hedman2018deep}. The selected scenes exhibit diverse capture styles, encompassing both confined indoor environments and expansive, unbounded outdoor settings.

\subsection{Results}

% \vspace{-0.8cm}
\begin{table*}[!t]
\centering
	\caption{Quantitative evaluation of our method compared to previous work across three datasets. The Mip-NeRF360~\cite{barron2022mip} dataset includes both indoor and outdoor scenarios. Tanks. and Deep. indicates the average results over the Tanks$\&$Templates~\cite{knapitsch2017tanks} and Deep Blending~\cite{hedman2018deep} datasets. Average represents the mean value across all scenarios in these datasets.}
% \vspace{2mm}
		\label{tab:comparison}
  % \vspace{-0.4cm}
	\small
	\scalebox{0.92}{
		\begin{tabular}{l|ccc|ccc|ccc}
			
			Dataset & \multicolumn{3}{c|}{Average} & \multicolumn{3}{c|}{Mip-NeRF360}  & \multicolumn{3}{c}{Tanks. and Deep. }   \\
			Method$|$Metric
			& PSNR$\uparrow$  & SSIM$\uparrow$     & LPIPS$\downarrow$  
			& PSNR$\uparrow$  & SSIM$\uparrow$     & LPIPS$\downarrow$  
			& PSNR$\uparrow$  & SSIM$\uparrow$     & LPIPS$\downarrow$     \\
			\hline 
			Plenoxels~\cite{fridovich2022plenoxels}& 22.77 & 0.666 & 0.457 & 23.08 & 0.625 & 0.463 & 22.07 & 0.757 & 0.444 \\
			INGP-Base~\cite{muller2022instant}& 24.49 & 0.699 & 0.373 & 25.30 & 0.671 & 0.371 & 22.67 & 0.760 & 0.377 \\
			INGP-Big~\cite{muller2022instant}& 24.93 & 0.724 & 0.336 & 25.59 & 0.699 & 0.331 & 23.44 & 0.781 & 0.347 \\
			M-NeRF360~\cite{barron2022mip}&
   \cellcolor{orange!40}27.11 & \cellcolor{yellow!40}0.803 & \cellcolor{yellow!40}0.241 & \cellcolor{red!40}27.69 & \cellcolor{yellow!40}0.792 & \cellcolor{yellow!40}0.237 & \cellcolor{yellow!40}25.81 & \cellcolor{yellow!40}0.830 & \cellcolor{yellow!40}0.251 \\
			3D-GS~\cite{kerbl20233d}& 
 \cellcolor{yellow!40}26.92 & \cellcolor{orange!40}0.832 & \cellcolor{orange!40}0.214 & \cellcolor{yellow!40}27.21 & \cellcolor{orange!40}0.815 & \cellcolor{orange!40}0.214 & \cellcolor{orange!40}26.27 & \cellcolor{red!40}0.872 & \cellcolor{red!40}0.213 \\
            Ours & 
\cellcolor{red!40}27.17 & \cellcolor{red!40}0.836 & \cellcolor{red!40}0.210 & \cellcolor{orange!40}27.48 & \cellcolor{red!40}0.821 & \cellcolor{red!40}0.209 & \cellcolor{red!40}26.44 & \cellcolor{red!40}0.872 & \cellcolor{orange!40}0.214
			
		\end{tabular}
	}
\end{table*}
% \vspace{-0.8cm}

% \vspace{-0.8cm}
\begin{figure}[!t]
\centering
\includegraphics[width=1\textwidth]{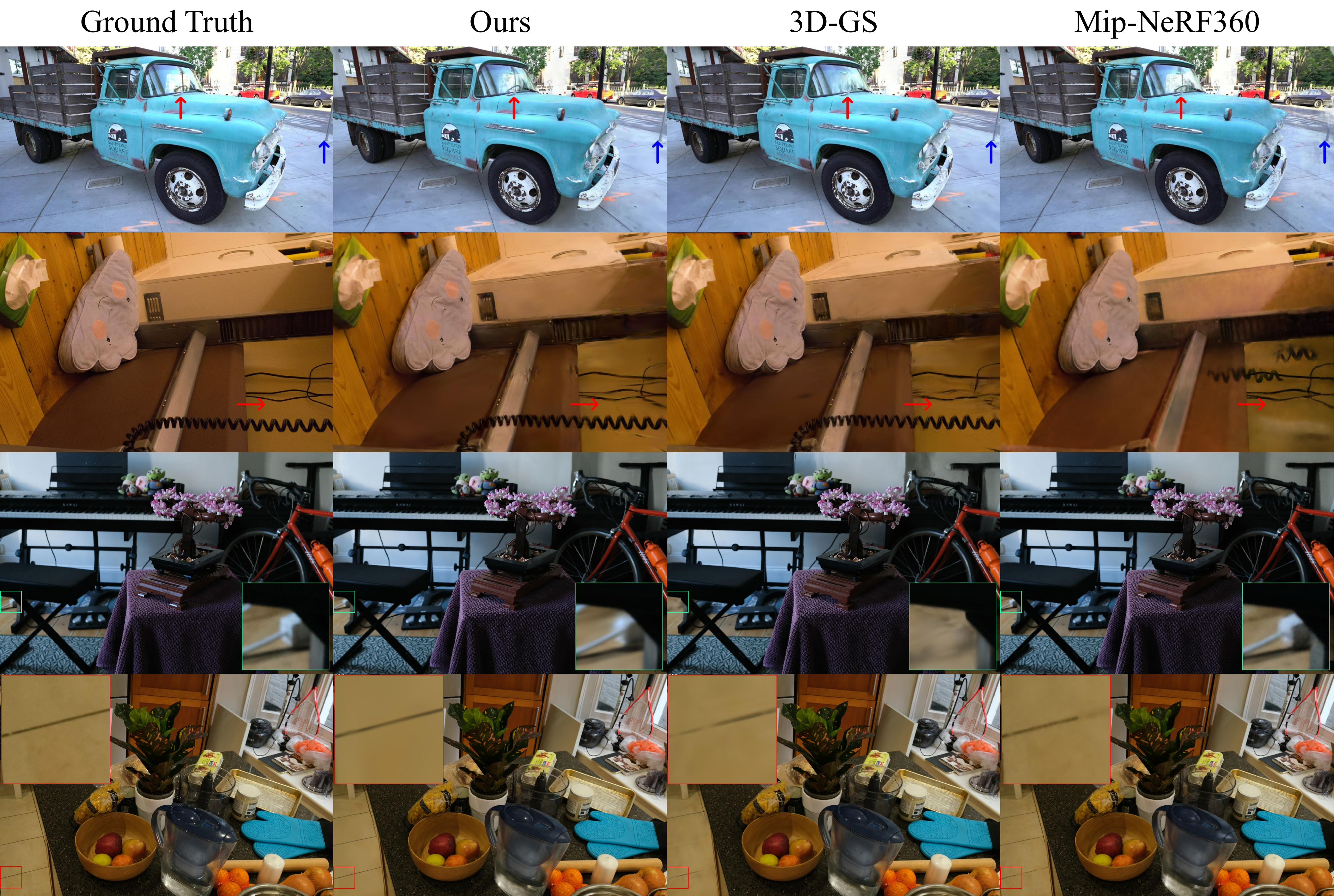}
\caption{We show comparisons of our method to previous methods and the corresponding ground truth images from held-out test views. The scenes are, from the top
down: \textsc{Truck} from Tanks$\&$Temples~\cite{knapitsch2017tanks}; 
\textsc{Playroom} from the Deep Blending dataset~\cite{hedman2018deep} and  \textsc{Bonsai}, \textsc{Counter} from Mip-NeRF360 dataset~\cite{barron2022mip}. Differences in quality highlighted by arrows/insets.}
\label{fig:comparison}
\vspace{-0.4cm}
\end{figure}

% \vspace{-0.8cm}
\begin{figure}[!t]
\centering
\includegraphics[width=1\textwidth]{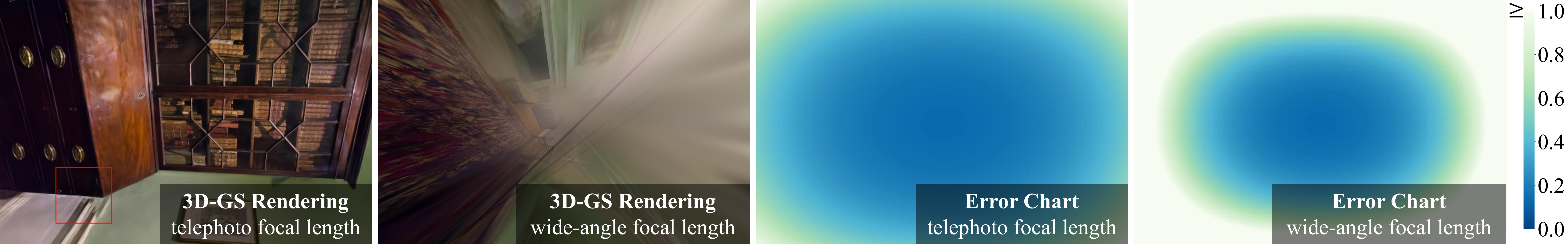}

\caption{We derive the projection error function with respect to image coordinates $(u, v)$ and focal length through the transformation between image and polar coordinates $(\theta_{\mu}, \phi_{\mu})$. We visualize this function, showing 3D-GS rendered images with long and short focal lengths, followed by the corresponding error functions.}
\label{fig:heatmap}
% \vspace{-0.4cm}
\end{figure}

We compare our algorithm with 3D-GS~\cite{kerbl20233d} and three state-of-the-art NeRF-based approaches: Mip-NeRF360~\cite{barron2022mip}, InstantNGP~\cite{muller2022instant} and Plenoxels~\cite{fridovich2022plenoxels}. All methods are configured using the same settings as outlined in the 3D-GS~\cite{kerbl20233d} to control variables, preventing the introduction of variables other than our Optimal Splatting method.

\paragraph{\textbf{Quantitative comparisons}}

We adopt a train/test split for datasets following the methodology proposed by Mip-NeRF360~\cite{barron2022mip}. Specifically, every 8th photo is reserved for testing, ensuring a consistent and meaningful basis for comparisons to generate error metrics. Standard metrics such as PSNR, LPIPS~\cite{lpips}, and SSIM are employed for evaluation. We report quantiative results in Table~\ref{tab:comparison}. While the 3D-GS significantly outperforms MipNeRF360 in rendering performance, its rendering quality slightly falls behind MipNeRF360, with a lower average PSNR across the 13 scenes. This might be attributed to MipNeRF360 not employing significant approximations throughout its entire pipeline. However, by replacing the original projection method of 3D-GS with our optimal projection, which incurs smaller errors, we not only maintain rendering performance far surpassing MipNeRF360 but also achieve a quality improvement.

\paragraph{\textbf{Qualitative comparisons}}

For our qualitative experiments, we select only two methods, 3D-GS and MipNeRF360, as our baselines. It can be observed that our method is capable of generating more realistic details, with fewer artifacts compared to 3D-GS~\cite{kerbl20233d}. 
And this is precisely the superiority brought about by our projection method, which results in smaller errors. Furthermore, We see that in some cases Mip-NeRF360
has remaining artifacts that our method avoids (e.g., black lines - in \textsc{Playroom}).

\subsection{Discussions}

\paragraph{\textbf{Impacts of Decreasing Focal Length}}
\label{para:large_fov}

Based on Equation~\ref{eq:error_function_thetaphi}, we further derive the projection error function with respect to image coordinates $(u, v)$ and focal length $f_x, f_y$ through the coordinate transformation between image coordinates and polar coordinates $(\theta_{\mu}, \phi_{\mu})$.
% \begin{equation}
% \left[\begin{matrix}t_1\\t_2\\t_3\end{matrix}\right]=
% \left[\begin{matrix}     \left(u - c_x\right)/f_x\\
%     \left(v - c_y\right)/f_y\\
%     1\end{matrix}\right]
% \end{equation}
% \begin{equation}
% \left[\begin{matrix}\theta_{\mu}\\\phi_{\mu}\end{matrix}\right]=\left[\begin{matrix}
%     \arctan\left(-t_2/\sqrt{t_1^2+t_3^2}\right)\\
%     \arctan\left(t_1/t_3\right)
% \end{matrix}\right]
% \end{equation}
As illustrated in Figure~\ref{fig:heatmap}, the peak of projection error mainly occurs at the image edges under a long focal length, resulting in artifacts at the image edges. As the focal length decreases, the field of view expands, leading to more Gaussians deviating from the projection center and an overall increase in projection error. In such cases, 3D-GS exhibit more artifacts, such as needle-like structures or cloud-like large Gaussians, which obscure parts that perform well under long focal length, thereby significantly degrading the overall image quality.
Since our method employs a central radial projection for the projection plane, theoretically, we should not encounter such defects. To validate this perspective, we conduct experiments at various focal length reduction ratios (reduction ratios of $\times0.2$ and $\times0.3$). 

\begin{table*}[!t]
\centering

	\caption{
We quantitatively compare our method with the original 3D-GS~\cite{kerbl20233d} using focal lengths of 0.2 and 0.3 times the original. Six scenes were selected from three datasets: Mip-NeRF360~\cite{barron2022mip}, Tank $\&$ Templates~\cite{knapitsch2017tanks}, and Deep Blending~\cite{hedman2018deep}. The scenes include \textsc{Train} and \textsc{Truck} from Tanks $\&$ Templates, \textsc{DrJohnson} and \textsc{Playroom} from Deep Blending, and \textsc{Room} and \textsc{Treehill} from Mip-NeRF360.}
		\label{tab:large_fov_comparison}
% \vspace{-0.2cm}
	\small
	\scalebox{0.96}{
		\begin{tabular}{c|c|c|cccccc|c}
Metric                 & Focal                & Method & Train          & Truck          & Dr Johnson      & Playroom       & Room           & Treehill       & Average        \\\hline
\multirow{4}{*}{PSNR$\uparrow$}  & \multirow{2}{*}{$\times$0.2} & 3D-GS~\cite{kerbl20233d}  & 10.12          & 12.39          & 19.24          & 16.38          & 20.02          & 15.36          & 15.58          \\
                       &                      & Ours   & \textbf{16.60} & \textbf{16.70} & \textbf{23.54} & \textbf{23.29} & \textbf{23.67} & \textbf{18.98} & \textbf{20.46} \\
                       & \multirow{2}{*}{$\times$0.3} & 3D-GS~\cite{kerbl20233d}  & 12.85          & 14.89          & 23.36          & 22.51          & 25.00          & 18.35          & 19.49          \\
                       &                      & Ours   & \textbf{17.52} & \textbf{18.07} & \textbf{24.89} & \textbf{26.40} & \textbf{25.87} & \textbf{19.87} & \textbf{22.10} \\\hline
\multirow{4}{*}{SSIM$\uparrow$}  & \multirow{2}{*}{$\times$0.2} & 3D-GS~\cite{kerbl20233d}  & 0.325          & 0.380          & 0.616          & 0.619          & 0.680          & 0.371          & 0.499          \\
                       &                      & Ours   & \textbf{0.565} & \textbf{0.537} & \textbf{0.701} & \textbf{0.752} & \textbf{0.777} & \textbf{0.435} & \textbf{0.628} \\
                       & \multirow{2}{*}{$\times$0.3} & 3D-GS~\cite{kerbl20233d}  & 0.471          & 0.529          & 0.740          & 0.791          & 0.812          & 0.437          & 0.630          \\
                       &                      & Ours   & \textbf{0.588} & \textbf{0.605} & \textbf{0.762} & \textbf{0.844} & \textbf{0.834} & \textbf{0.472} & \textbf{0.684} \\\hline
\multirow{4}{*}{LPIPS$\downarrow$} & \multirow{2}{*}{$\times$0.2} & 3D-GS~\cite{kerbl20233d}  & 0.569          & 0.479          & 0.296          & 0.355          & 0.207          & 0.474          & 0.397          \\
                       &                      & Ours   & \textbf{0.277} & \textbf{0.250} & \textbf{0.216} & \textbf{0.203} & \textbf{0.178} & \textbf{0.385} & \textbf{0.251} \\
                       & \multirow{2}{*}{$\times$0.3} & 3D-GS~\cite{kerbl20233d}  & 0.331          & 0.276          & 0.207          & 0.213          & \textbf{0.154} & 0.397          & 0.263          \\
                       &                      & Ours   & \textbf{0.243} & \textbf{0.217} & \textbf{0.201} & \textbf{0.182} & 0.158          & \textbf{0.385} & \textbf{0.231}
		\end{tabular}
	}
\end{table*}

% \vspace{-0.4cm}
\begin{figure}[!t]
\centering
% \vspace{-0.2cm}
\includegraphics[width=1\textwidth]{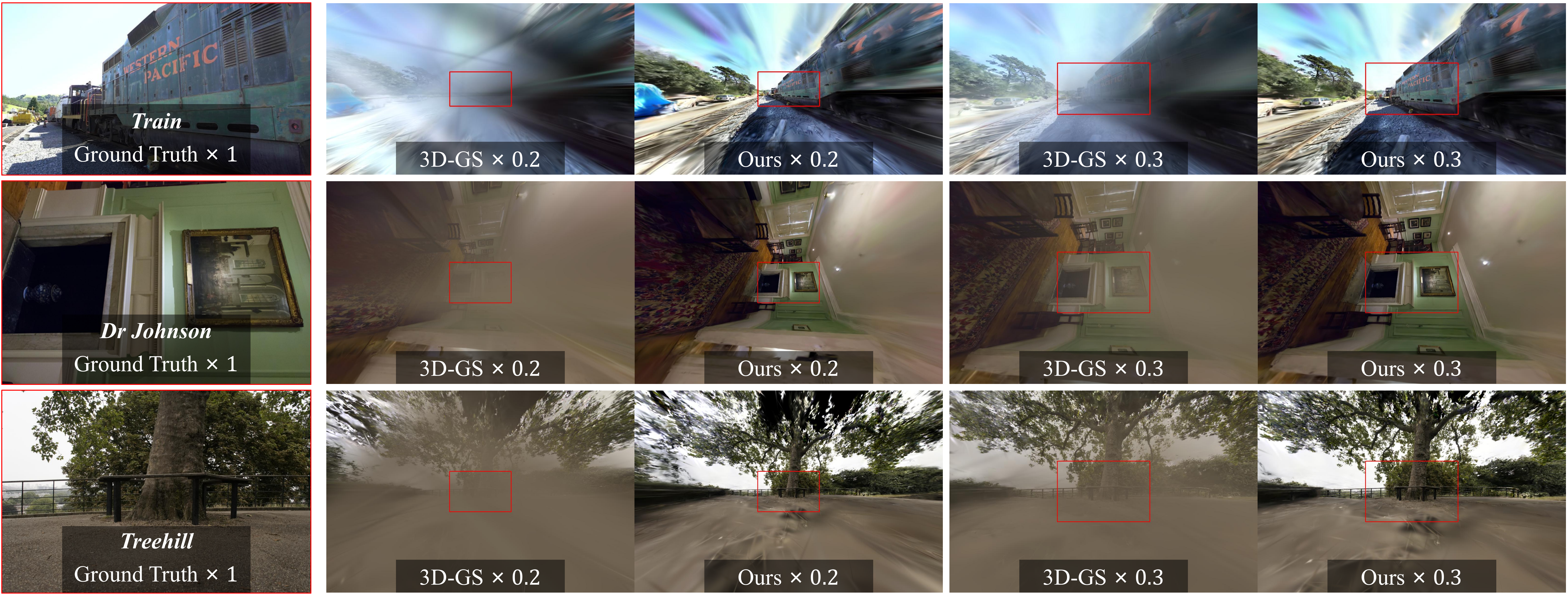}
\caption{We compare our method to the original 3D-GS~\cite{kerbl20233d} with different focal length reductions. Two focal lengths, 0.2 and 0.3 times the original, are tested. Ground truth $\times1$ in the left images indicates no focal length reduction. The red-boxed areas in the right images correspond to the same regions in the left images.}
\label{fig:large_fov_comparison}
% \vspace{-0.4cm}
\end{figure}

% \vspace{-0.4cm}
\begin{figure}[!htbp]
\centering
\includegraphics[width=1\textwidth]{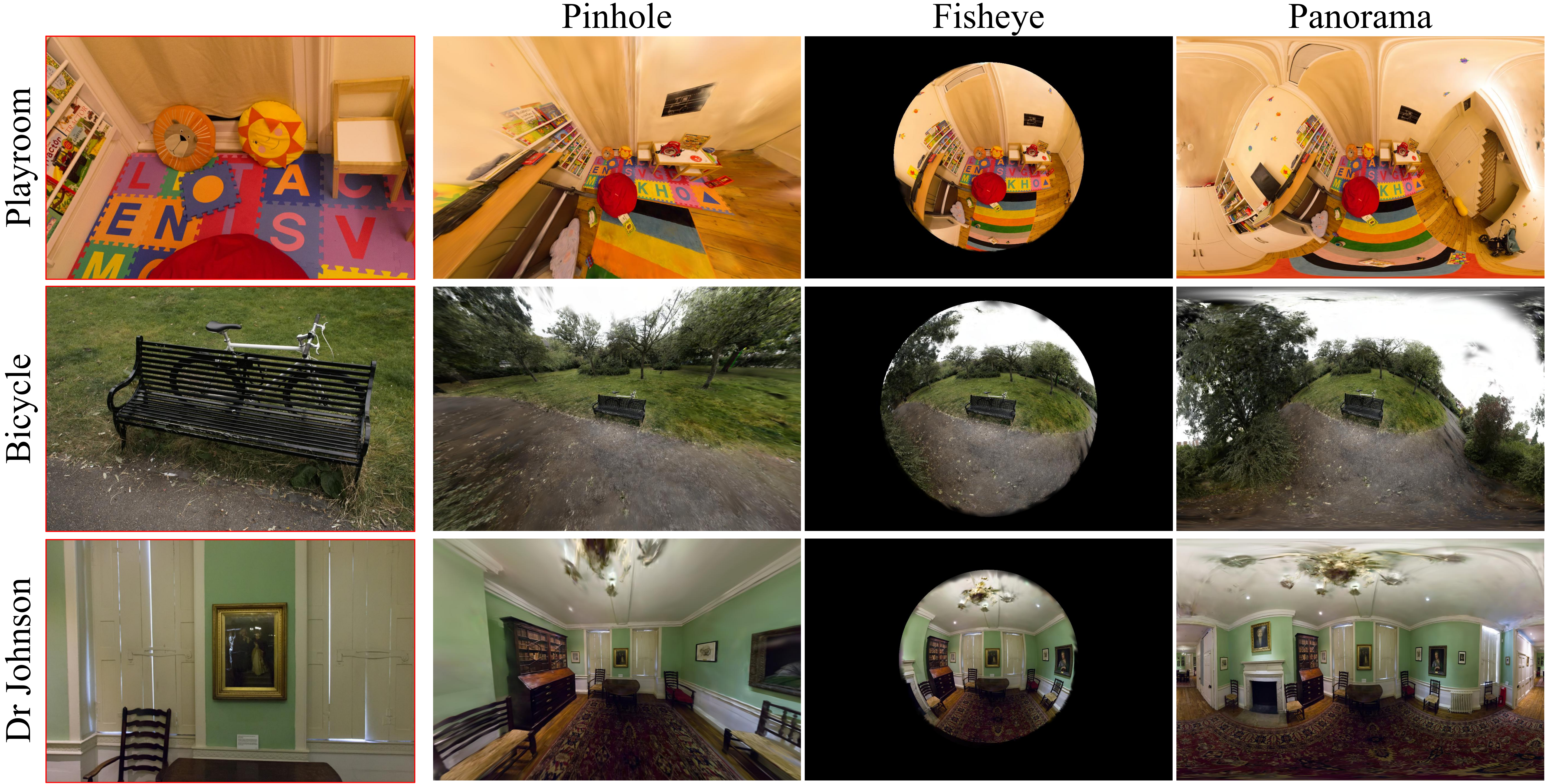}
\caption{Renderings of three scenes using our projection under three camera models. From left to right: the ground truth of the original perspective, wide-angle rendering with a pinhole camera, fisheye camera rendering, and the panorama.}
\label{fig:fisheye}
% \vspace{-0.4cm}
\end{figure}

For quantitative experiments, considering that there is no ground truth for large field-of-view in the 13 scenes from the 3D-GS~\cite{kerbl20233d}, we apply a focal length mask. This mask selects central parts of the rendered images with corresponding ground truth based on the focal length scaling factor for metric calculation. The results are presented in the Table \ref{tab:large_fov_comparison}. Whether it is at $\times0.2$ or $\times0.3$ reduction ratios, our method outperforms the original 3D-GS significantly across various scenes in terms of all metrics. As the focal length decreases from $\times0.3$ to $\times0.2$, 3D-GS~\cite{kerbl20233d} experiences a significant degradation due to increased projection errors, while our method demonstrates greater robustness.

For quantitative experiments, we randomly select three scenes, each with two different focal lengths for demonstration. Additional experimental results can be found in the supplementary materials. It is evident from the Figure~\ref{fig:large_fov_comparison} that our method exhibits greater robustness across various focal length settings, mitigating the artifacts introduced by projection in the original 3D-GS.

\paragraph{\textbf{Adaptability to Various Camera Models}}

By straightforwardly modifying the transformation from image space to camera space, we have achieved adaptation to various camera models, with the caveat that the original 3D-GS's projection struggles to support these camera models. Figure~\ref{fig:fisheye} illustrates that our projection, in addition to conforming to the original pinhole camera according to Equation~\ref{eq:img2tan}, can also accommodate fisheye cameras and generate panoramas.

\paragraph{\textbf{Limitation}}

Due to our method's independence from the image plane, it exhibits adaptability to various camera models. However, the slight increase in training time is a consequence of the subsequent transformation to the image plane after projection onto the tangent plane, rather than a direct projection onto the image plane. Fortunately, this method avoids the need for dense points sampling, ensuring the algorithm's time complexity remains unchanged and rendering performance stays unaffected. Moreover, it may be possible to mitigate the impact on training time, particularly concerning the corresponding backward propagation, through the implementation of optimized CUDA code.

\section{Conclusion}

Through theoretical analysis, we establish the correlation between the projection error and the Gaussian mean in 3D-GS. By optimizing the error function, we discover that the projection error is minimized when projecting on the plane perpendicular to the line connecting the Gaussian mean and the camera center. Leveraging this insight, we propose a novel projection that yields high-quality images without compromising rendering performance and can be easily adapted to various camera models through simple modifications.

In this work, our primary focus is on analyzing the relationship between the projection error and the mean of the 3D Gaussian under the assumption of a constant covariance. The influence of Gaussian covariance on projection requires further discussion in future work. It may be possible to further explore the potential of Gaussian Splatting as an explicit representation technique for scenes.

% \clearpage
% \section*{Acknowledgements}
% Please insert your acknowledgments here.

\bibliographystyle{splncs04}
\bibliography{main}

% ---- Appendix ----

 \clearpage

\appendix

\section{Details of Error Analysis}

\paragraph{\textbf{Unit Vector Simplification}}

We have proved that the composition of the transformation $\varpi$, projecting onto the unit sphere, and the transformation $\varphi$, projecting onto the tangent plane, is equivalent to the single transformation $\varphi$ directly projecting onto the tangent plane:

\begin{equation}
\begin{split}
\left(\varphi\circ\varpi\right)\left({\mathbf{x}^{'}}\right)
&=\varphi\left({\mathbf{x}^{'}}{\left({\mathbf{x}^{'}}^{\top}{\mathbf{x}^{'}}\right)^{-1/2}}\right)\\
&={\mathbf{x}^{'}}{\left({\mathbf{x}^{'}}^{\top}{\mathbf{x}^{'}}\right)^{-1/2}}\left(\mathbf{x_0}^{\top}\left({\mathbf{x}^{'}}{\left({\mathbf{x}^{'}}^{\top}{\mathbf{x}^{'}}\right)^{-1/2}}\right)\right)^{-1}\\
&={\mathbf{x}^{'}} \left(\mathbf{x_0}^{\top}{\mathbf{x}^{'}}\right)^{-1}\\
&=\varphi\left({\mathbf{x}^{'}}\right)\text{.}
\end{split}
\end{equation}
Therefore, we can simplify the derivation of the error function $\epsilon$ by projecting all relevant points onto the unit sphere as unit vectors (Equation 8 in the main paper).

\paragraph{\textbf{Closed-Form Expression for the Error Function}}

This section provides a comprehensive derivation of the integral expression for the error function as presented in the main paper.  Upon expressing $\mathbf{x}^{'}$, $\mathbf{x_0}$ and $\boldsymbol{\mu}^{'}$ as unit vectors and substituting them into Equation 6 in the main paper, the Jacobian matrix is derived:

\begin{equation}
\mathbf{J}=\frac{\partial \varphi}{\partial \mathbf{x}^{'}}\left(\boldsymbol{\mu}^{'}\right)=
\left[\begin{matrix}\frac{1}{\cos{\left(\phi_{\mu} \right)} \cos{\left(\theta_{\mu} \right)}} & 0 & - \frac{\sin{\left(\phi_{\mu} \right)}}{\cos^{2}{\left(\phi_{\mu} \right)} \cos{\left(\theta_{\mu} \right)}}\\0 & \frac{1}{\cos{\left(\phi_{\mu} \right)} \cos{\left(\theta_{\mu} \right)}} & \frac{\sin{\left(\theta_{\mu} \right)}}{\cos^{2}{\left(\phi_{\mu} \right)} \cos^{2}{\left(\theta_{\mu} \right)}}\\0 & 0 & 0\end{matrix}\right]\text{.}
\end{equation} Substituting this Jacobian matrix into Equation 5 in the main paper yields the derivation of the Taylor expansion remainder term:

\begin{equation}
R_1\left(\mathbf{x}^{'}\right)
=\left[\begin{matrix}- \frac{\sin{\left(\phi - \phi_{\mu} \right)} \cos{\left(\theta \right)}}{\cos^{2}{\left(\phi_{\mu} \right)} \cos{\left(\theta_{\mu} \right)}} + \tan{\left(\phi \right)} - \tan{\left(\phi_{\mu} \right)}\\\frac{\sin{\left(\theta \right)}}{\cos{\left(\phi_{\mu} \right)} \cos{\left(\theta_{\mu} \right)}} - \frac{\sin{\left(\theta_{\mu} \right)} \cos{\left(\phi \right)} \cos{\left(\theta \right)}}{\cos^{2}{\left(\phi_{\mu} \right)} \cos^{2}{\left(\theta_{\mu} \right)}} + \frac{\tan{\left(\theta_{\mu} \right)}}{\cos{\left(\phi_{\mu} \right)}} - \frac{\tan{\left(\theta \right)}}{\cos{\left(\phi \right)}}\\0\end{matrix}\right]\text{.}
\end{equation} And substituting this expression into Equation 7 in the main paper results in the integral expression (Equation 9 in the main paper). The main paper provides the integral expression (Equation 9 in the main paper) and the function graph (Figure 2 in the main paper) for the error function, without presenting the closed-form expression after integration. We provide the closed-form expression as follows:

\begin{equation}
\begin{split}
\epsilon&\left(\theta_{\mu}, \phi_{\mu}\right)=\frac{1}{{16 \cos^{4}{\left(\phi_{\mu} \right)} \cos^{4}{\left(\theta_{\mu} \right)} \tan{\left(\phi_{\mu} + \frac{\pi}{4} \right)} \tan{\left(\theta_{\mu} + \frac{\pi}{4} \right)}}}\\&\Big\{8 \left(\left(2 \tan{\left(\theta_{\mu} + \frac{\pi}{4} \right)} - \pi\right) \tan{\left(\theta_{\mu} + \frac{\pi}{4} \right)} + 2\right) \cos^{4}{\left(\phi_{\mu} \right)} \cos^{4}{\left(\theta_{\mu} \right)} \tan^{2}{\left(\phi_{\mu} + \frac{\pi}{4} \right)} +\\&\left. \left(- 4 \cdot \left(2 \cos{\left(2 \theta_{\mu} \right)} + \pi\right) \sin^{2}{\left(\phi_{\mu} \right)} + \pi \left(2 \cos{\left(2 \theta_{\mu} \right)} + \pi\right) + 4 \cos{\left(2 \theta_{\mu} \right)} + 2 \pi\right) \right.\\&\left.\sin^{2}{\left(\theta_{\mu} \right)} \tan{\left(\phi_{\mu} + \frac{\pi}{4} \right)} \tan{\left(\theta_{\mu} + \frac{\pi}{4} \right)} + 8\log{\frac{{\left(\sin{\left(\phi_{\mu} + \frac{\pi}{4} \right)} + 1 \right)}{\left( \cos{\left(\phi_{\mu} + \frac{\pi}{4} \right)} + 1 \right)}}{{\left(1-\sin{\left(\phi_{\mu} + \frac{\pi}{4} \right)} \right)} {\left(1 - \cos{\left(\phi_{\mu} + \frac{\pi}{4} \right)} \right)}}}\right.\\&\left.\log{\frac{{\left(1-\sin{\left(\phi_{\mu} + \frac{\pi}{4} \right)} \right)} {\left(1 - \cos{\left(\phi_{\mu} + \frac{\pi}{4} \right)} \right)} {\exp{\left(2 \sqrt{2} \cos{\left(\theta_{\mu} \right)}\right)}}}{
{\left(\sin{\left(\phi_{\mu} + \frac{\pi}{4} \right)} + 1 \right)}{\left( \cos{\left(\phi_{\mu} + \frac{\pi}{4} \right)} + 1 \right)}{\left(\tan^{2\sin{\left(\theta_{\mu} \right)}}{\left(\theta_{\mu} + \frac{\pi}{4} \right)} \right)}}}\right.\\&\left. \cos^{3}{\left(\phi_{\mu} \right)} \cos^{3}{\left(\theta_{\mu} \right)} \tan{\left(\phi_{\mu} + \frac{\pi}{4} \right)} \tan{\left(\theta_{\mu} + \frac{\pi}{4} \right)} +\right.\\&\left. 4 \pi \left(- 4 \log{\left(\tan{\left(\phi_{\mu} + \frac{\pi}{4} \right)} \right)} \tan{\left(\phi_{\mu} \right)} + \pi \tan^{2}{\left(\phi_{\mu} \right)} + 2 \tan{\left(\phi_{\mu} + \frac{\pi}{4} \right)} - \pi\right) \right.\\&\left.\cos^{4}{\left(\phi_{\mu} \right)} \cos^{4}{\left(\theta_{\mu} \right)} \tan{\left(\phi_{\mu} + \frac{\pi}{4} \right)} \tan{\left(\theta_{\mu} + \frac{\pi}{4} \right)} + \right.\\&\left.\left(2 \pi \left(\left(4+2 \pi+16 \sqrt{2}\right) \sin^{2}{\left(\theta_{\mu} \right)} - 2 + \pi\right) \cos^{2}{\left(\phi_{\mu} \right)} + \left(2 \pi- 4\right) \cos{\left(2 \theta_{\mu} \right)}   - 2 \pi + \pi^{2}\right)\right.\\&\left. \cos^{2}{\left(\theta_{\mu} \right)} \tan{\left(\phi_{\mu} + \frac{\pi}{4} \right)} \tan{\left(\theta_{\mu} + \frac{\pi}{4} \right)} + \right.\\&\left.16 \sqrt{2} \log{\frac{{\left(1-\sin{\left(\phi_{\mu} + \frac{\pi}{4} \right)} \right)} {\left(1 - \cos{\left(\phi_{\mu} + \frac{\pi}{4} \right)} \right)} {\exp{\left(2 \sqrt{2} \cos{\left(\theta_{\mu} \right)}\right)}}}{
{\left(\sin{\left(\phi_{\mu} + \frac{\pi}{4} \right)} + 1 \right)}{\left( \cos{\left(\phi_{\mu} + \frac{\pi}{4} \right)} + 1 \right)}}}\right.\\&\left. \cos^{3}{\left(\phi_{\mu} \right)} \cos^{4}{\left(\theta_{\mu} \right)} \tan{\left(\phi_{\mu} + \frac{\pi}{4} \right)} \tan{\left(\theta_{\mu} + \frac{\pi}{4} \right)} +\right.\\&\left. 64 \sin^{2}{\left(\phi_{\mu} \right)} \cos^{2}{\left(\phi_{\mu} \right)} \cos^{4}{\left(\theta_{\mu} \right)} \tan{\left(\phi_{\mu} + \frac{\pi}{4} \right)} \tan{\left(\theta_{\mu} + \frac{\pi}{4} \right)} -\right.\\&\left. 16 \sqrt{2} \cdot \left(1 + \sqrt{2}\right) \sin{\left(\theta_{\mu} \right)} \sin{\left(2 \theta_{\mu} \right)} \cos^{2}{\left(\phi_{\mu} \right)} \cos{\left(\theta_{\mu} \right)} \tan{\left(\phi_{\mu} + \frac{\pi}{4} \right)} \tan{\left(\theta_{\mu} + \frac{\pi}{4} \right)} +\right.\\& 8 \pi \cos^{4}{\left(\phi_{\mu} \right)} \cos^{4}{\left(\theta_{\mu} \right)} \tan{\left(\theta_{\mu} + \frac{\pi}{4} \right)} + \frac{8 \left(- 2 \pi \cos^{2}{\left(\theta_{\mu} \right)} + \pi + 4\right) \cos^{4}{\left(\phi_{\mu} \right)} \cos^{2}{\left(\theta_{\mu} \right)}}{\left(\tan{\left(\theta_{\mu} \right)} - 1\right)^{2}}\Big\}\text{.}
\end{split}
\end{equation}

\section{Details of Adaptation for Various Camera Models}

In the main paper, it is highlighted that our projection, independent of perspective image plane, allows for the adaptation to various camera models by modifying the transformation from image space to camera space in the rasterization based on the unit sphere (Equation 16 in the main paper) . This section provides a detailed explanation of our method's adaptation to various camera models.

\paragraph{\textbf{Fisheye}}

The design models for fisheye cameras can generally be categorized into four types: equidistant projection model, equisolid angle projection model, orthographic projection model, and stereographic projection model. We explane our projection's adaptation to fisheye camera models, taking the equidistant projection model as an example.

For a pixel $\left(u, v\right)$ on the image, we cast a ray. According to the transformation between the image space and camera space for the equidistant projection model, we obtain:

\begin{equation}
\mathbf{x}_{\text{2D}}=\varphi_{\text{p}}\left(\left[\begin{matrix} \frac{\left(u-c_{x}\right) \sin{\left(\sqrt{\frac{\left(u - c_{x}\right)^{2}}{f_{x}^{2}}+\frac{\left(v-c_{y}\right)^{2}}{f_{y}^{2}}} \right)}}{f_{x} \sqrt{\frac{\left(u - c_{x}\right)^{2}}{f_{x}^{2}}+\frac{\left(v-c_{y}\right)^{2}}{f_{y}^{2}}} }\\\frac{\left(v-c_{y}\right) \sin{\left(\sqrt{\frac{\left(u - c_{x}\right)^{2}}{f_{x}^{2}}+\frac{\left(v-c_{y}\right)^{2}}{f_{y}^{2}}} \right)}}{f_{y} \sqrt{\frac{\left(u - c_{x}\right)^{2}}{f_{x}^{2}}+\frac{\left(v-c_{y}\right)^{2}}{f_{y}^{2}}} }\\\cos{\left(\sqrt{\frac{\left(u - c_{x}\right)^{2}}{f_{x}^{2}}+\frac{\left(v-c_{y}\right)^{2}}{f_{y}^{2}}} \right)}\end{matrix}\right]\right)
\end{equation} where $c_x$, $c_y$, $f_x$, $f_y$ denote the intrinsic parameters of the camera model.

\paragraph{\textbf{Panorama}}

For a pixel $\left(u, v\right)$ on the image, we cast a ray. According to the transformation between the image space and camera space for panoramic images, we obtain:
\begin{equation}
\mathbf{x}_{\text{2D}}=\varphi_{\text{p}}\left(\left[\begin{matrix}\sin{\left(\frac{\pi \left(- W + 2 u\right)}{W} \right)} \cos{\left(\frac{\pi \left(- \frac{H}{2} + v\right)}{H} \right)}\\\sin{\left(\frac{\pi \left(- \frac{H}{2} + v\right)}{H} \right)}\\\cos{\left(\frac{\pi \left(- \frac{H}{2} + v\right)}{H} \right)} \cos{\left(\frac{\pi \left(- W + 2 u\right)}{W} \right)}\end{matrix}\right]\right)
\end{equation} where $H, W$ represent the height and width of the image, respectively.

\paragraph{}

Then, similar to Equation 17 in the main paper, the 2D Gaussian function values can be obtained for alpha blending to generate the image. The alpha blending process after obtaining the 2D Gaussian is the same as in the original 3D-GS~\cite{kerbl20233d}.

% \vspace{-0.4cm}
\begin{figure}[!t]
\centering
\includegraphics[width=1\textwidth]{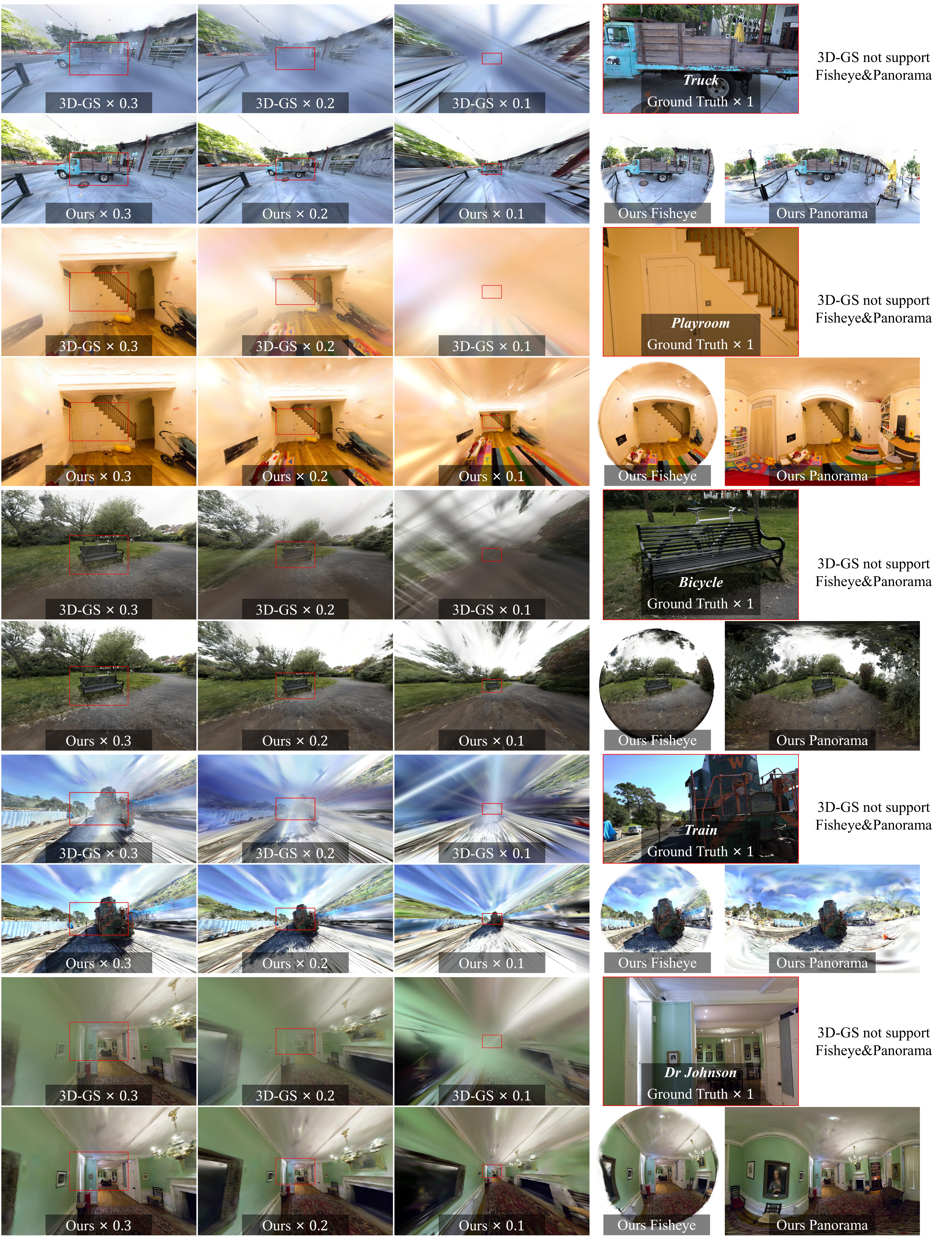}
\caption{We show comparisons of our method to the original 3D-GS~\cite{kerbl20233d} under various camera models and different focal lengths. }
\label{fig:large_fov_comparison4supple}
% \vspace{-0.8cm}
\end{figure}

% Three focal lengths under the pinhole camera model are provided, 0.3, 0.2 and 0.1 times the original test set focal length. In the right images, the Ground truth $\times1$ represents the Ground truth without focal length reduction. In the left images, the regions enclosed by the red box correspond to the areas in the Ground truth. We also illustrate the renderings under our fisheye camera model and panoramas, which the original 3D-GS can not produce.

\section{Details of Experiments}

\paragraph{\textbf{Additional Results}}

\begin{figure}[ht]
\centering
\includegraphics[width=1\textwidth]{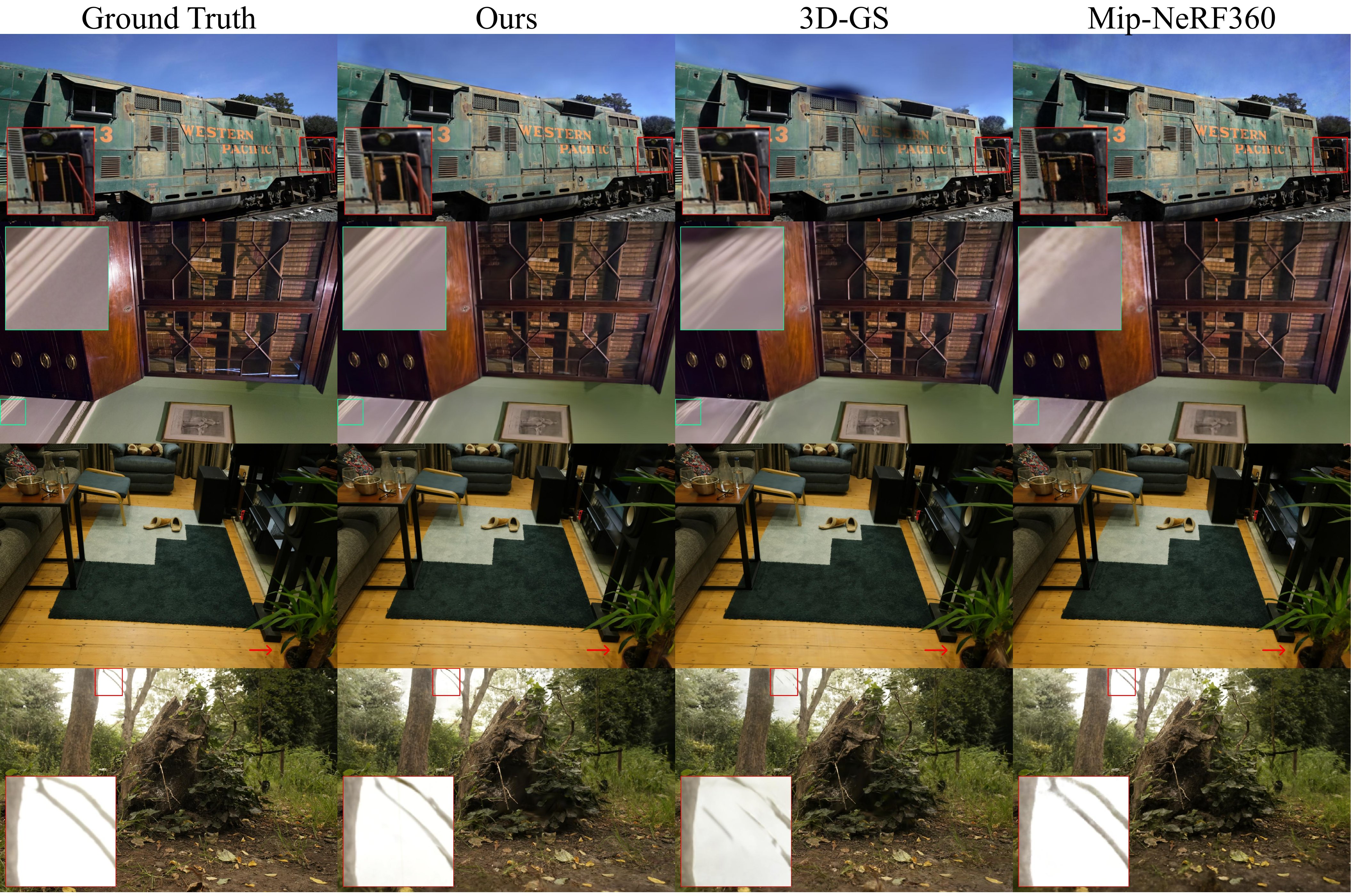}
\caption{We show comparisons of our method to previous methods and the corresponding ground truth images from held-out test views. The scenes are, from the top
down: \textsc{Train} from Tanks\&Temples~\cite{knapitsch2017tanks}; 
\textsc{Drjohnson} from the Deep Blending dataset~\cite{hedman2018deep} and  \textsc{Room}, \textsc{Stump} from Mip-NeRF360 dataset~\cite{barron2022mip}. Differences in quality highlighted by arrows/insets.}
\label{fig:comparison4supple}
% \vspace{-0.4cm}
\end{figure}

Table~\ref{tab:360_scene_psnr}-\ref{tab:ttdb_scene_psnr} list PSNR for our evaluation over all considered techniques and real-world scenes, corresponding to Table 1 in the main paper. Figure~\ref{fig:comparison4supple} illustrates additional qualitative comparisons between our method and other approaches.

% \vspace{-0.8cm}
% \vspace{-0.8cm}

\paragraph{\textbf{Impacts of Decreasing Focal Length}}

% \vspace{-0.8cm}
\begin{figure}[!htbp]
\centering
\includegraphics[width=1\textwidth]{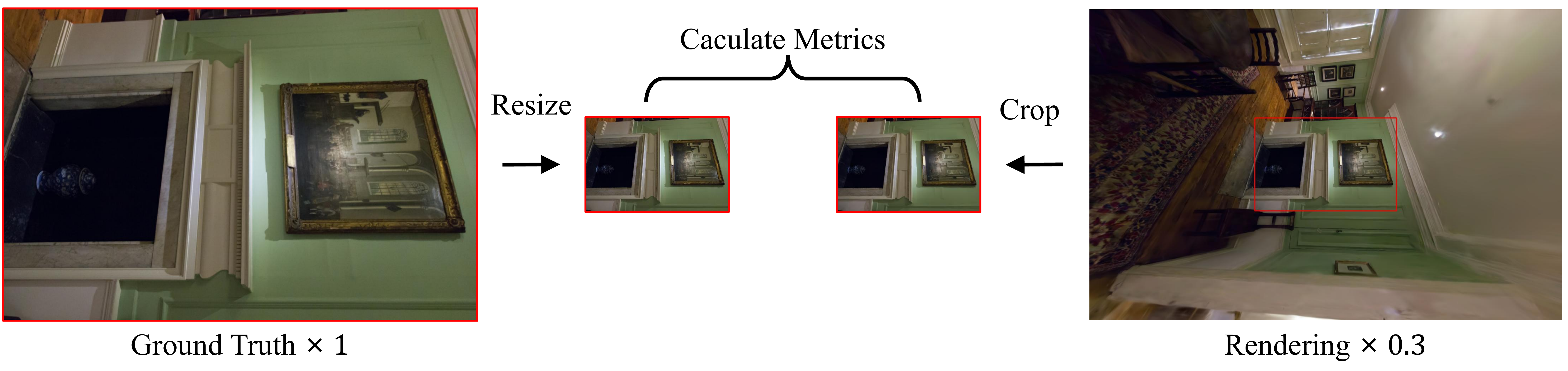}
\caption{Illustration of details of metrics caculation in the quantitative comparisons on the impact of decreasing focal length.}
\label{fig:metrics4large_fov}
% \vspace{-0.4cm}
\end{figure}

In the quantitative comparison experiments about the impact of reducing focal length, we utilized a focal length mask methodology due to the absence of ground truth for wide-angle images. The detailed procedures are illustrated in Figure~\ref{fig:metrics4large_fov}. The focal mask selects central parts of the rendering with corresponding ground truth based on the focal length scaling factor for metric calculation.
For rendering, we crop the corresponding part with available ground truth, then resize the ground truth to match the size of this patch. Finally, we compute the metrics for these two images.

For qualitative experiments, we show additional results in Figure~\ref{fig:large_fov_comparison4supple}.
In addition to the original focal length reduction factors of 0.3 and 0.2, we introduce 0.1 times the focal length. Additionally, we generate images with corresponding perspectives using other camera models, while 3D-GS~\cite{kerbl20233d} is not presented as it lacks support for these camera models.

\paragraph{\textbf{Comparison with the Other 3D-GS Methods}}

\begin{figure}[!htbp]
\centering
\includegraphics[width=1\textwidth]{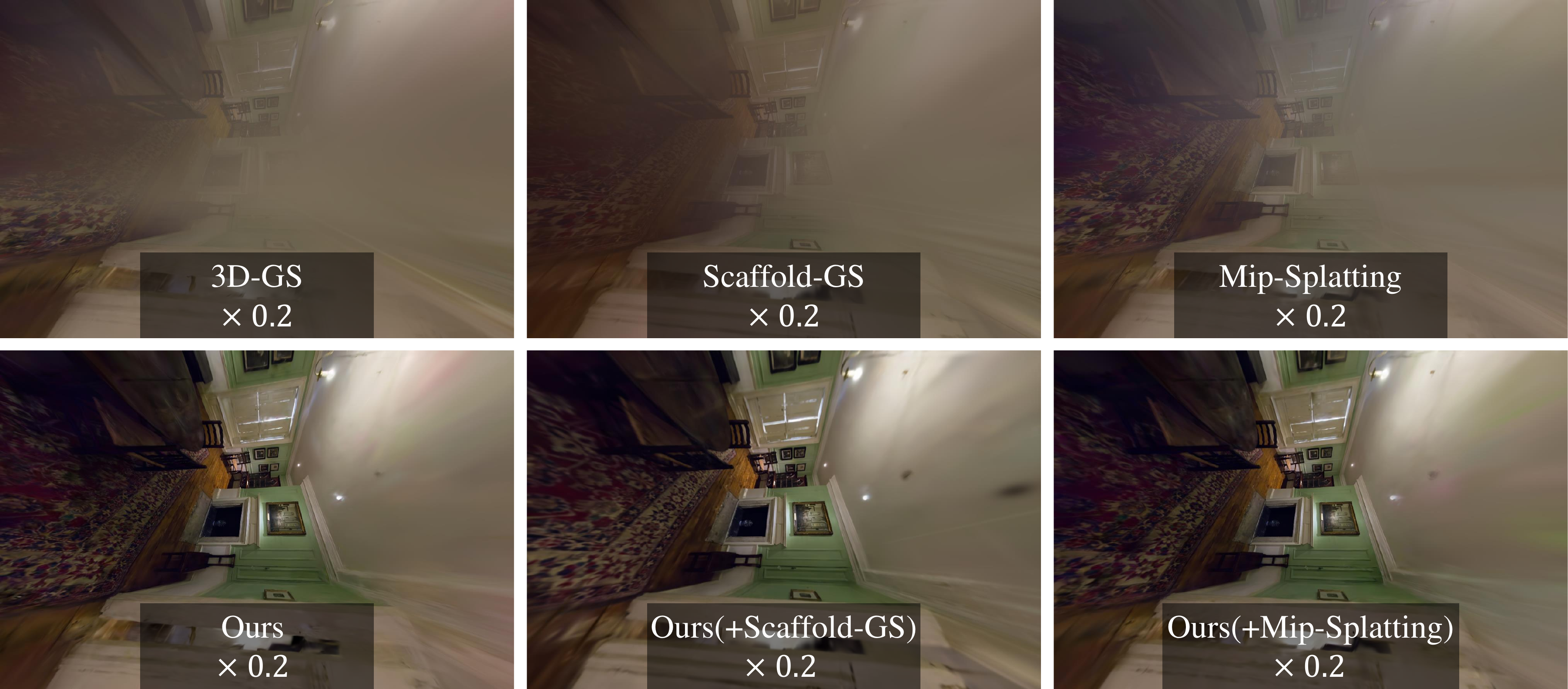}
\caption{We show comparisons of our method to the original 3D-GS~\cite{kerbl20233d}, Mip-Splatting~\cite{mip_gs} and Scaffold-GS~\cite{scaffold} with a large FOV camera. In the figure, Ours(+X) indicates the combination of our method with the X method.}
\label{fig:large_fov_comparison_oth_gs}
\end{figure}

We choose Mip-Splatting~\cite{mip_gs} and Scaffold-GS~\cite{scaffold} to compare in the same setting as in Figure 6 of the main paper.
For Mip-Splatting~\cite{mip_gs} and Scaffold-GS~\cite{scaffold} still employ conventional projection of 3D-GS, the approximation error remains larger compared to ours, as shown in Figure~\ref{fig:large_fov_comparison_oth_gs}. Additionally, our method can be be seamlessly integrated with theirs to reduce the projection error of these methods.

\paragraph{\textbf{Adaptability to Various Camera Models}}

\begin{figure}[!htbp]
\centering
\includegraphics[width=1\textwidth]{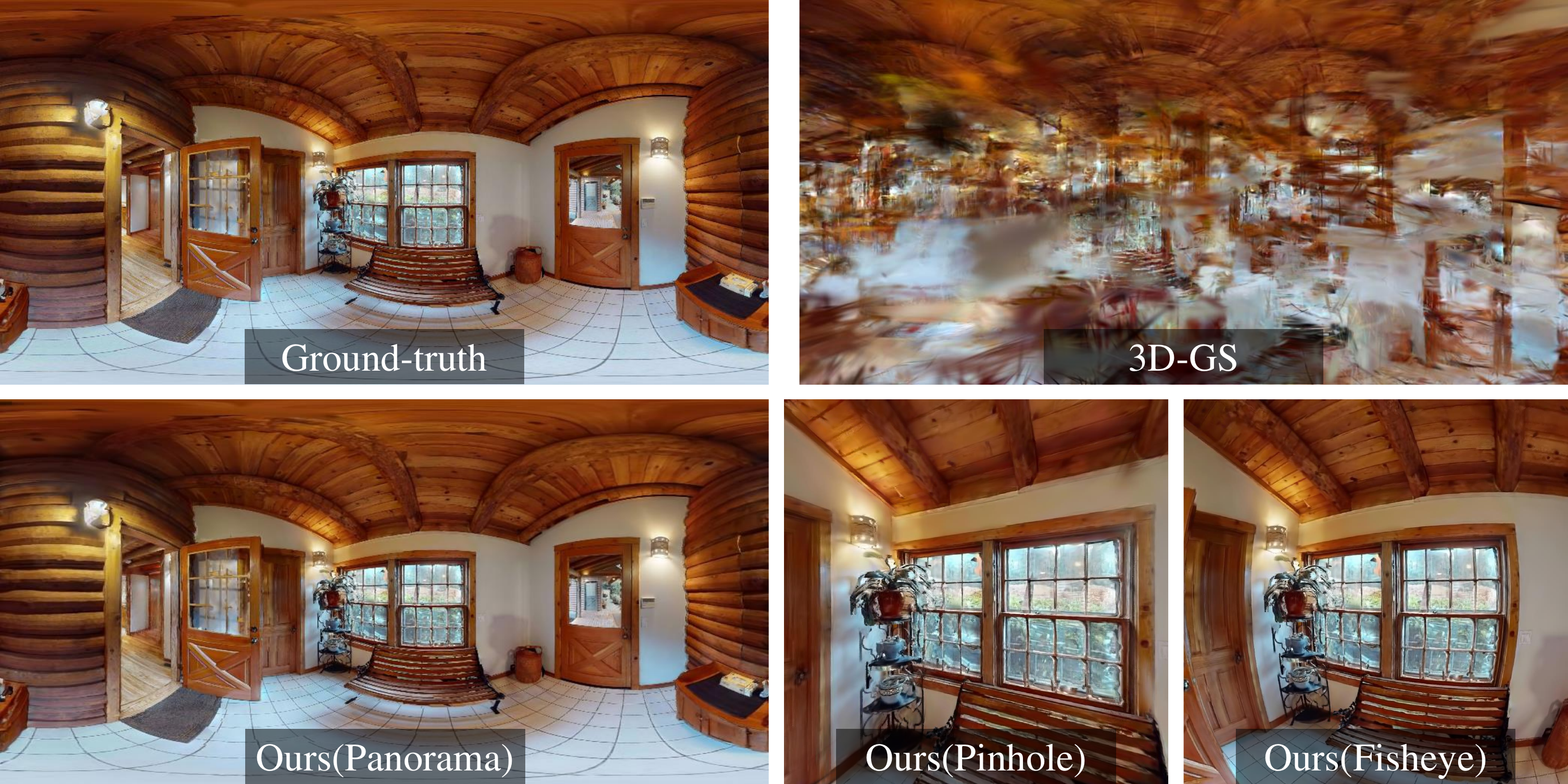}
\caption{We show the results of our method trained on the non-pinhole camera dataset \textsc{Matterport}~\cite{chang2017matterport3d}.}
\label{fig:train_360}
\end{figure}

Since our method is based on differentiable rasterizer, it can also be used to train on the non-pinhole camera dataset. We illustrate the results obtained from training on the non-pinhole camera dataset \textsc{Matterport}~\cite{chang2017matterport3d} directly, where 3D-GS fails entirely to reconstruct the scene, as shown in Figure~\ref{fig:train_360}.

% \vspace{-0.4cm}
\begin{table}[ht]
		\centering
	\caption{PSNR scores for Mip-NeRF360 scenes. }
        % \vspace{2mm}
	\scalebox{0.78}{
		\centering
\begin{tabular}{l|ccccc|cccccc|c}
&\multicolumn{5}{c|}{MipNeRF360 Indoor}      & \multicolumn{6}{c|}{MipNeRF360 Outdoor}                & \multicolumn{1}{c}{\multirow{2}{*}{Avg.}} \\
&Room  & Counter & Kitchen & Bonsai & Avg.  & Bicycle & Flowers & Garden & Stump & Treehill & Avg.  &                \\
Plenoxels~\cite{fridovich2022plenoxels} & 27.59 & 23.62   & 23.42   & 24.67  & 24.83 & 21.91   & 20.10   & 23.49  & 20.66 & 22.25    & 21.68 & 23.08                                     \\
INGP-Base~\cite{muller2022instant}& 29.27 & 26.44   & 28.55   & 30.34  & 28.65 & 22.19   & 20.35   & 24.60  & 23.63 & 22.36    & 22.63 & 25.30                                     \\
INGP-Big~\cite{muller2022instant}& 29.69 & 26.69   & 29.48   & 30.69  & 29.14 & 22.17   & 20.65   & 25.07  & 23.47 & 22.37    & 22.75 & 25.59                                     \\
M-NeRF360~\cite{barron2022mip}&\cellcolor{red!40}31.63 & \cellcolor{red!40}29.55   & \cellcolor{red!40}32.23   & \cellcolor{red!40}33.46  & \cellcolor{red!40}31.72 & \cellcolor{yellow!40}24.37   & \cellcolor{red!40}21.73   & \cellcolor{yellow!40}26.98  & \cellcolor{yellow!40}26.40 & \cellcolor{red!40}22.87    & \cellcolor{yellow!40}24.47 & \cellcolor{red!40}27.69                                     \\
3D-GS~\cite{kerbl20233d}& \cellcolor{yellow!40}30.63 & \cellcolor{yellow!40}28.70   & \cellcolor{yellow!40}30.32   & \cellcolor{yellow!40}31.98  & \cellcolor{yellow!40}30.41 & \cellcolor{red!40}25.25   & \cellcolor{yellow!40}21.52   & \cellcolor{red!40}27.41  & \cellcolor{orange!40}26.55 & \cellcolor{yellow!40}22.49    & \cellcolor{red!40}24.64 & \cellcolor{yellow!40}27.21                                     \\
Ours & \cellcolor{orange!40}31.58 & \cellcolor{orange!40}29.05   & \cellcolor{red!40}31.23   & \cellcolor{orange!40}32.31  & \cellcolor{orange!40}31.04 & \cellcolor{orange!40}25.07   & \cellcolor{orange!40}21.54   & \cellcolor{orange!40}27.15  & \cellcolor{red!40}26.57 & \cellcolor{orange!40}22.77    & \cellcolor{orange!40}24.62 & \cellcolor{orange!40}27.48                                    
\end{tabular}
	}
	\label{tab:360_scene_psnr}
 % \vspace{-0.4cm}
\end{table}

	\begin{table}[ht]
 		\centering
	\caption{PSNR scores for Tanks\&Temples and Deep Blending scenes. }
 % \vspace{2mm}
	\scalebox{0.78}{
		\centering
\begin{tabular}{l|ccc|ccc|c}
&\multicolumn{3}{c|}{Tanks\&Temples}      & \multicolumn{3}{c|}{Deep Blending}                & \multicolumn{1}{c}{\multirow{2}{*}{Avg.}} \\
~ & Truck & Train & Avg. & Dr Johnson & Playroom & Avg. & ~                \\
Plenoxels~\cite{fridovich2022plenoxels} & 
23.22 & 18.93 & 21.07 & 23.14 & 22.98 & 23.06 & 22.07\\
INGP-Base~\cite{muller2022instant}&
23.26 & 20.17 & 21.72 & 27.75 & 19.48 & 23.62 & 22.67\\
INGP-Big~\cite{muller2022instant}&
23.38 & \cellcolor{yellow!40}20.46 & 21.92 & 28.26 & 21.67 & 24.96 & 23.44\\
M-NeRF360~\cite{barron2022mip}&
\cellcolor{yellow!40}24.91 & 19.52 & \cellcolor{yellow!40}22.22 & \cellcolor{red!40}29.14 & \cellcolor{yellow!40}29.66 & \cellcolor{yellow!40}29.40 & \cellcolor{yellow!40}25.81\\
3D-GS~\cite{kerbl20233d}&
\cellcolor{red!40}25.19 & \cellcolor{orange!40}21.10 & \cellcolor{orange!40}23.14 & \cellcolor{yellow!40}28.77 & \cellcolor{red!40}30.04 & \cellcolor{orange!40}29.41 & \cellcolor{orange!40}26.27 \\
Ours &
\cellcolor{red!40}25.19 & \cellcolor{red!40}21.67 & \cellcolor{red!40}23.43 & \cellcolor{red!40}29.14 & \cellcolor{orange!40}29.89 & \cellcolor{red!40}29.51 & \cellcolor{red!40}26.47  
\end{tabular}
	}
\label{tab:ttdb_scene_psnr}
 % \vspace{-0.4cm}
\end{table}

\end{document}